\documentclass[pdflatex,sn-mathphys-num]{sn-jnl}
\usepackage{graphicx}%
\usepackage{multirow}%
\usepackage{amsmath,amssymb,amsfonts}%
\usepackage{amsthm}%
\usepackage{mathrsfs}%
\usepackage[title]{appendix}%
\usepackage{xcolor}%
\usepackage{textcomp}%
\usepackage{manyfoot}%
\usepackage{booktabs}%
\usepackage{algorithm}%
\usepackage{algorithmicx}%
\usepackage{algpseudocode}%
\usepackage{listings}%
\usepackage{longtable} 
\usepackage{placeins}
\usepackage{comment}
\usepackage{soul}
\usepackage{chngcntr}
\usepackage{caption}
\usepackage{url}

\newcommand{\llmClaude}{Anthropic Claude 3 Opus}
\newcommand{\llmGpt}{OpenAI GPT-4}
\newcommand{\llmMistral}{Mistral Large}
\newcommand{\llmGemini}{Google Gemini Pro 1.0}
\newcommand{\llmLlama}{Meta Llama 2 70B}
\usepackage[modulo]{lineno}

\unnumbered%

\begin{document}

\title{Human-AI collectives produce the most accurate differential diagnoses}

\author*[1]{\fnm{Nikolas} \sur{Z{\"o}ller}}\email{zoeller@mpib-berlin.mpg.de}
\author[1]{\fnm{Julian} \sur{Berger}}
\author[2]{\fnm{Irving} \sur{Lin}}
\author[2]{\fnm{Nathan} \sur{Fu}}
\author[2]{\fnm{Jayanth} \sur{Komarneni}}
\author[3]{\fnm{Gioele} \sur{Barabucci}}
\author[2]{\fnm{Kyle} \sur{Laskowski}}
\author[4]{\fnm{Victor} \sur{Shia}}
\author[5]{\fnm{Benjamin} \sur{Harack}}
\author[6]{\fnm{Eugene A.} \sur{Chu}}
\author[7]{\fnm{Vito} \sur{Trianni}}
\author*[1,8]{\fnm{Ralf H.J.M.} \sur{Kurvers}}\email{kurvers@mpib-berlin.mpg.de}
\author*[1]{\fnm{Stefan M.} \sur{Herzog}}\email{herzog@mpib-berlin.mpg.de}


\affil*[1]{\orgdiv{Center for Adaptive Rationality}, \orgname{Max Planck Institute for Human Development}, \orgaddress{\state{Berlin}, \country{Germany}}}
\affil[2]{\orgname{The Human Diagnosis Project}, \orgaddress{\city{San Francisco} , \state{CA}, \country{USA}}}
\affil[3]{\orgname{University of Cologne}, \orgaddress{\city{Cologne},\country{Germany}}}
\affil[4]{\orgname{Harvey Mudd College}, \orgaddress{\city{Claremont}, \state{CA}, \country{USA}}}
\affil[5]{\orgname{Oxford University}, \orgaddress{\city{Oxford}, \country{UK}}}
\affil[6]{\orgname{Kaiser Permanente}, \orgaddress{\city{Downey}, \state{CA}, \country{USA}}}
\affil[7]{\orgname{ISTC-CNR}, \orgaddress{\city{Roma}, \country{Italy}}}
\affil[8]{
\orgdiv{Science of Intelligence Excellence Cluster}, \orgname{ Technical University Berlin}, \orgaddress{ \city{Berlin}, \country{Germany}}}

\abstract{
Artificial intelligence systems, particularly large language models (LLMs), are increasingly being employed in high-stakes decisions that impact both individuals and society at large, often without adequate safeguards to ensure safety, quality, and equity.
Yet LLMs hallucinate \cite{hong2024hallucinations,Ji2023,Pal2023,tonmoy2024Comprehensive}, lack common sense \cite{williams2024easy}, and are biased \cite{Omiye2023,Navigli2023}---shortcomings that may reflect LLMs' inherent limitations and thus may not be remedied by more sophisticated architectures, more data, or more human feedback. Relying solely on LLMs for complex, high-stakes decisions is therefore problematic.
Here we present a hybrid collective intelligence system that mitigates these risks by leveraging the complementary strengths of human experience and the vast information processed by LLMs.
We apply our method to open-ended medical diagnostics, combining 40,762 differential diagnoses made by physicians with the diagnoses of five state-of-the art LLMs across 2,133 medical cases. We show that hybrid collectives of physicians and LLMs outperform both single physicians and physician collectives, as well as single LLMs and LLM ensembles. This result holds across a range of medical specialties and professional experience, and can be attributed to humans' and LLMs' complementary contributions that lead to different kinds of errors.
Our approach highlights the potential for collective human and machine intelligence to improve accuracy in complex, open-ended domains \cite{Trianni:HHAI2023} like medical diagnostics.
}


\maketitle


Diagnostic errors are among the most pressing issues in medical practice \cite{makary2016medical, leape1991nature, graber2005diagnostic}, causing an estimated 795,000 deaths and permanent disabilities in the United States alone each year \cite{newman2024}. Reducing diagnostic errors---without incurring substantially higher costs---is essential to improve patient outcomes worldwide.
This challenge has motivated a recent surge in diagnostic technologies exploiting artificial intelligence (AI) to interpret medical records, tests, and images \cite{basu2020artificial,mirbabaie2021artificial}.
Deep learning approaches in medical imaging have shown great promise. Notable examples include mammography interpretation, cardiac function assessment, and lung cancer screening, some of which have progressed beyond the testing phase and entered clinical practice \cite{rajpurkar2022ai, aggarwal2021diagnostic, dembrowerArtificialIntelligenceBreast2023}. 

Recent years have also witnessed the rise of AI foundation models, especially LLMs, which show remarkable abilities to process natural language, providing accurate answers to questions in almost any domain, including medicine \cite{lu2024multimodal,moorFoundationModelsGeneralist2023,singhalLargeLanguageModels2023,Jiang2023}. However, a recent meta-analysis \cite{takita2024diagnostic} found that physicians often outperform LLMs, and that LLMs differ vastly in performance, also between medical specialties.
While LLMs' performance in the medical domain keeps improving \cite{singhalLargeLanguageModels2023}, their deployment in clinical practice remains challenging due to the risk of errors (caused by, e.g., hallucinations \cite{hong2024hallucinations,Ji2023,Pal2023,tonmoy2024Comprehensive}, biases \cite{Omiye2023,Navigli2023}, and lack of common sense \cite{williams2024easy}) and concerns about their trustworthiness \cite{chakravortiAITrustProblem2024}. As these shortcomings may reflect inherent limitations of LLMs \cite{lenat2023getting}, developing more sophisticated architectures or using more data or more human feedback may not sufficiently address these shortcomings.
The tension between the vast potential of AI-based solutions and the challenges of real-world deployment is not limited to medical diagnostics. It is also apparent in other domains, especially those involving high-stakes decisions whose effects are not immediate, such as strategies to address climate change \cite{Cowls-ClimateAIGambit-2023}.

Here we present an approach that complements AI responses with human expert knowledge in open-ended medical diagnostics. This method, which combines AI with a collective intelligence (CI) approach, benefits from the diversity of solutions provided by humans and LLMs. CI approaches harness the contributions of multiple experts to reduce errors and find creative solutions to complex problems \cite{Woolley:kh2010,Woolley-2024}. In medical diagnostics, several studies have found that the collective solution of multiple diagnosticians outperforms the average individual across a range of medical contexts \cite{hasan2024boosting,hautz2015diagnostic,kattan2016wisdom, kurvers2016boosting,kammer2017potential,kurvers2018combining,Blanchard2024}.
These studies have focused on binary or small-scale decision problems (e.g., detecting a specific condition), but CI has also proved successful in open-ended medical problems. While in earlier studies the contributions of individual experts are manually harmonized and aggregated into collective diagnoses \cite{barnett2019comparative}, more recently this approach has been fully automatized. Specifically, medical knowledge graphs and natural language processing methods are leveraged to harmonize the free-text contributions of individual experts \cite{kurvers2023}, which can differ significantly due to the open-endedness of the solution space \cite{Trianni:HHAI2023}. 

In a similar vein, hybrid systems that integrate state-of-the-art LLMs as peers in a mixed human--AI collective hold promise for addressing complex decision problems such as medical diagnostics. AI can provide complementary information without perpetuating the errors and biases of human peers. At the same time, the diagnostic process is not entirely outsourced to artificial systems, making it possible to benefit from human experts' ability to think outside the box, recognize context, and handle contentious evidence, thus mitigating the risks of LLMs.

Combining the contributions of multiple humans and multiple LLMs is, however, not straightforward. Although many studies have explored how to combine multiple AI models (e.g., ensemble learning is an established practice in machine learning \cite{kuncheva2014combining,mienye2022survey}), little is known about how to best combine the outputs of multiple LLMs (but see \cite{jiang2023llm,jiang2024mixtral,yang2023one,barabucci2024combining} for specific use cases), or how to combine the responses of multiple LLMs with those of human experts, particularly in open-ended domains. In this study, we develop a general-purpose method to combine the responses produced by both human experts and LLMs. 
Applying this method to a set of over 40,000 diagnoses, we show that hybrid human--AI collectives outperform human-only and LLM-only collectives across a variety of medical specialties and levels of professional experience. Additionally, we demonstrate that when LLMs fail, physicians often provide correct diagnoses, thus highlighting the crucial importance of maintaining expert involvement, even in the presence of an ensemble of powerful AIs.

\section{Medical cases, human data and LLM responses}

The empirical basis for this work is a dataset from the Human Diagnosis Project (Human Dx), an online collaborative platform for medical professionals and trainees. Users from around the world can register on the platform, submit cases, review case details, and provide diagnoses. The cases submitted are published only if approved by an editorial board of licensed medical professionals. Each case is presented as a vignette mimicking information that physicians encounter in real-world practice and containing patient information such as symptoms, medical records, and clinical test results (see Fig. \ref{fig:schema}). When responding to a case, users can provide either a single diagnosis or a ranked list, commonly known as a differential diagnosis,  either as free text or by selecting from a medical taxonomy with an auto-complete feature that activates as they type (see Fig. \ref{fig:schema}a for an illustration of the user interface). We refer to this response as a differential diagnosis, whether it contains one or multiple diagnoses. Once the user has submitted their differential diagnosis, they are shown the solution as provided by cases' authors and vetted by an expert panel, which may consist of one or several diagnoses.
For our main analyses, we used a set of 2,133 medical cases and 40,762 differential diagnoses from qualified physicians with different levels of professional experience (see Methods). In the Supplement, we additionally present results of the same analyses for medical students. 

To compare and combine the human diagnoses with LLM outputs, we provided the same set of case vignettes to five commercially available or open-source state-of-the-art LLMs (\llmClaude, \llmGemini, \llmLlama, \llmMistral, and \llmGpt) and prompted the models to provide the five most probable diagnoses, ordered by their probability of being correct (see Methods).

\section{Harmonizing and aggregating open-ended answers from doctors and LLMs}

\begin{figure}[t!]
\centering
\includegraphics[width=1.0\textwidth]{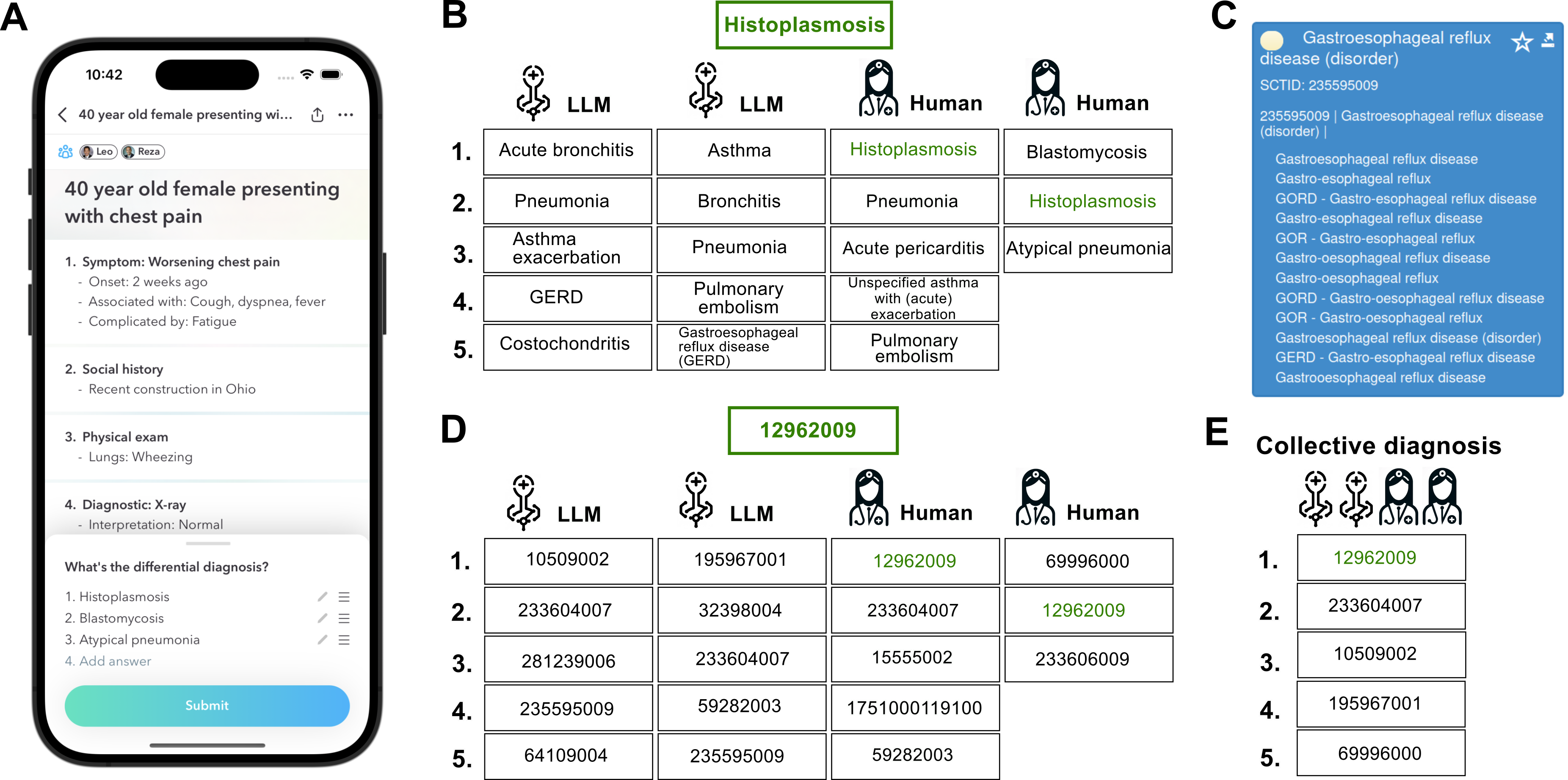}
\caption[Illustration of the hybrid human--AI collective intelligence process]{\textbf{Illustration of the hybrid collective intelligence process, which combines human diagnoses with LLM outputs to arrive at a collective differential diagnosis.} \textbf{a}, Screenshot of the interface that human users see when diagnosing a patient case on the Human Dx platform via a mobile device. The information provided can include a patient's symptoms, test results, and medical record. Users can uncover this information piece by piece and update their diagnosis accordingly. In this analysis, we only consider users' final differential diagnosis. The same information shown to human users is also given to LLMs as part of a prompt (see Methods). \textbf{b}, An illustrative example of the open-ended text responses given by users and LLMs. Next, extending a method presented in \cite{kurvers2023} (see Methods and Extended Data Fig.~\ref{fig:prompt_engineering}), each single diagnosis is subjected to several processing steps for standardization, after which it is assigned a unique ID in the SNOMED CT healthcare terminology. \textbf{c}, Example of a SNOMED CT entry. Crucially, all listed synonyms are matched to the same SNOMED CT ID. \textbf{d}, Diagnoses of humans and LLMs after the matching step. \textbf{e}, Collective diagnosis after aggregating the diagnoses from humans and LLMs. In this aggregation, LLMs and humans are assigned different weights based on their performance in the training fold. The rank $r$ of a diagnosis in a differential diagnosis is taken into account through a $1/r$ scoring rule (see Methods).}\label{fig:schema}
\end{figure}

The procedure for harmonizing and aggregating the human judgements and the LLM output is illustrated in Fig.~\ref{fig:schema}. In short, a collective diagnosis is established by ranking each diagnosis with a weighted score that is calculated across all humans and LLMs in the group, taking into account both the rank of the diagnosis in the individual lists (i.e., higher-ranked diagnoses receive more weight) and the humans' and LLMs' performance on a training fold (see Methods).
We used a five-fold cross-validation approach. One fold was used as a training fold to select the best prompting configuration for each LLM and to calculate weights for each LLM and each human using a method proposed by \cite{Dogan2019}. All humans were assigned the same weight based on the average human performance on the cases in the training fold. The collective diagnoses were then evaluated on the remaining four folds (see Methods, Extended Data Fig. \ref{fig:prompt_engineering}, and SI for details).

In order to make the open-ended diagnoses of users and LLMs comparable and uniquely identifiable, we extended the method described in \cite{kurvers2023}, which maps free-text diagnoses to concepts (and their unique IDs) in the Systematized Nomenclature of Medicine Clinical Terms (SNOMED CT) \cite{donnelly2006snomed} (see Methods). SNOMED CT is a comprehensive clinical terminology and coding system designed to standardize the representation of medical concepts and support the accurate communication of clinical information in healthcare. After matching diagnoses to SNOMED-CT concepts, the generation of the collective differential diagnoses proceeds exploiting the SNOMED-CT IDs (see Fig. \ref{fig:schema}).

\section{Aggregating LLMs increases performance in open-ended medical diagnostics}

\begin{figure}[t!]
\centering
\includegraphics[width=1.0\textwidth]{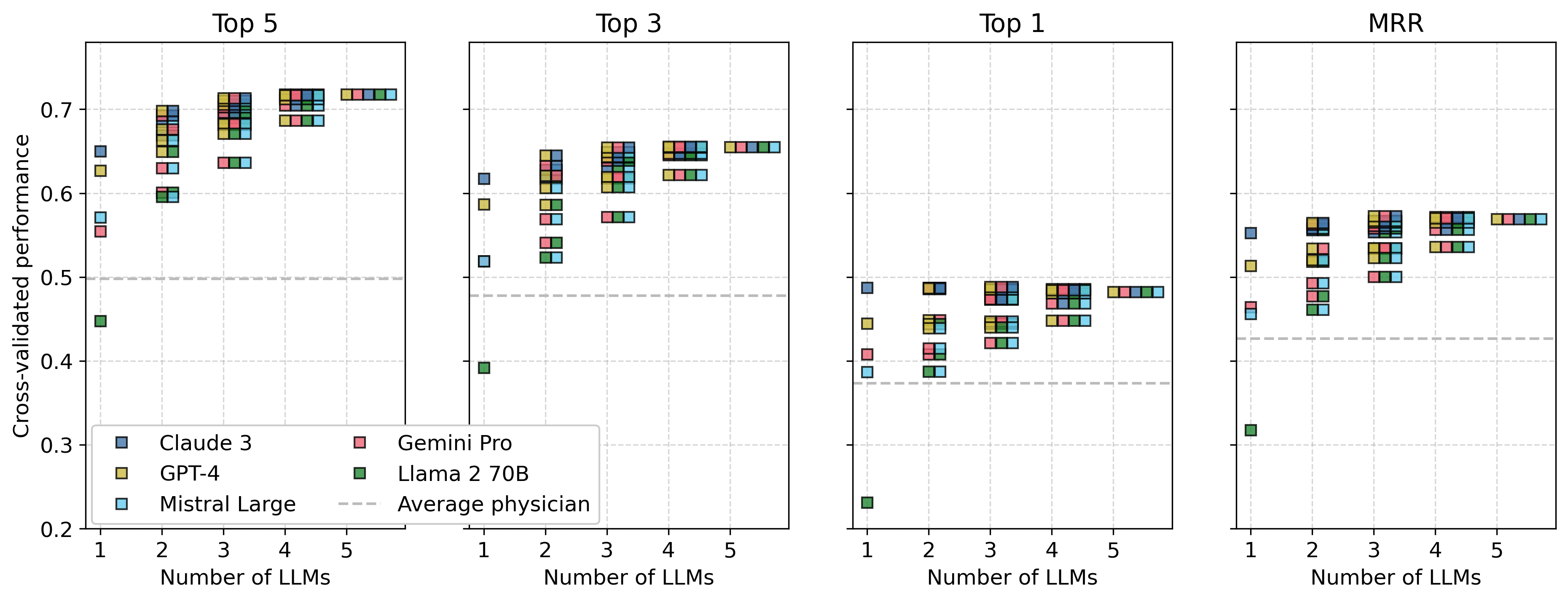}
\caption[Cross-validated performance of five individual LLMs and ensembles of all possible combinations of LLMs]{\textbf{Cross-validated performance of five individual LLMs (\llmClaude, \llmGpt, \llmMistral, \llmGemini\, and \llmLlama) and ensembles of all possible combinations of LLMs}.
Panels show performance for four outcome metrics ($y$ axes): Top-$k$ indicates the proportion of cases for which the correct diagnosis was among the $k$ top-ranked diagnoses (for $k = \{1,3, 5\}$); MRR shows the mean reciprocal rank of correct diagnoses across cases (see eq. \ref{eq:MRR}).
The $x$ axis shows the number of LLMs in an ensemble.
The horizontal dashed line shows the average individual performance of the physicians (i.e., first averaged within cases, then across all cases). Some of the ensembles overplot each other (see Table~\ref{tbl:LLMs} in the supplement for the performance of all combinations).}\label{fig:acc-llm-ensembles}
\end{figure}

We start by presenting the cross-validated results for the baseline performance of the five individual LLMs and all possible LLM ensembles.
Depending on the use case for an aggregated collective diagnosis, some performance metrics might be more suitable than others. For example, if the differential diagnosis of an LLM (ensemble), a human collective, or a hybrid collective serves as a consideration set to support the decision of a human physician, it may be sufficient that the correct solution is included in the differential diagnosis at all, and less important that it is ranked first. Therefore, we report several accuracy metrics, including top-$5$, top-$3$, and top-$1$ accuracy, where a differential diagnosis is evaluated as correct if the correct diagnosis is among the top five, top three, or top one diagnoses, respectively (and the accuracy is the proportion of such cases). For the fraction of cases where a case author has stated several diagnoses as correct ($34\%$), a nominated diagnosis is considered correct if it matches any of the correct diagnoses.
Additionally, we report the mean reciprocal rank (MRR) \cite{voorhees1999trec}, a well-established performance metric in the field of information retrieval, defined as
\begin{equation}\label{eq:MRR}
\text{MRR} = \frac{1}{C} \sum_{i=1}^{C} \frac{1}{\text{r}_i},
\end{equation}
where $C$ corresponds to the number of cases on which the metric is evaluated and $r_i$ is the rank of the correct answer in the final list for case $i$. Note that if $r_i > 5$ or if the correct diagnosis is not present in the ranking, we set $r_i = \infty$ so that the contribution of case $i$ to the MRR is null.

As Fig. \ref{fig:acc-llm-ensembles} shows, the individual LLMs differed notably in performance, but aggregating multiple LLMs into ensembles generally increased diagnostic accuracy. The ensembles performed much better than the worst individual LLM and generally as well as, or better than, the best individual LLM.
For top-$5$ accuracy, the ensemble of all LLMs combined clearly outperformed each LLM individually, and this result held across the five most common medical specialties in our data (cardiology, gastroenterology, pulmonology and respirology, neurology, and infectious diseases; see Extended Data Fig.~\ref{fig:LLMs_specialty}). The same held for top-$3$ accuracy and MRR when comparing performance across all cases, and for four of the five medical specialties (see Extended Data Fig. \ref{fig:LLMs_specialty}). For top-$1$ accuracy, the ensemble of all LLMs performed better than four of the five individual LLMs, but slightly worse than the best-performing LLM.
Whether or not it is advisable to aggregate several LLMs may therefore depend on the target metric, but if the purpose is to provide a consideration set to support the decision of a human physician (e.g., top-$5$ diagnoses), then LLM ensembles have the greatest potential.

To put this performance into perspective, Extended Data Fig. \ref{fig:performance_more} shows the percentage of physicians who were outperformed by (and/or tied with) individual LLMs and LLM ensembles across the set of cases they had solved. This percentage was highest for an LLM ensemble incorporating all five LLMs (i.e., strictly outperformed $85\%$ of physicians and outperformed or tied with $93\%$ of physicians).
Comparing the individual LLM performance with that of the human users showed that four of the five LLMs outperformed the average physician.

\section{Human--AI collective intelligence outperforms both humans and LLMs}

\begin{figure}[b]
\centering
\includegraphics[width=1\textwidth]{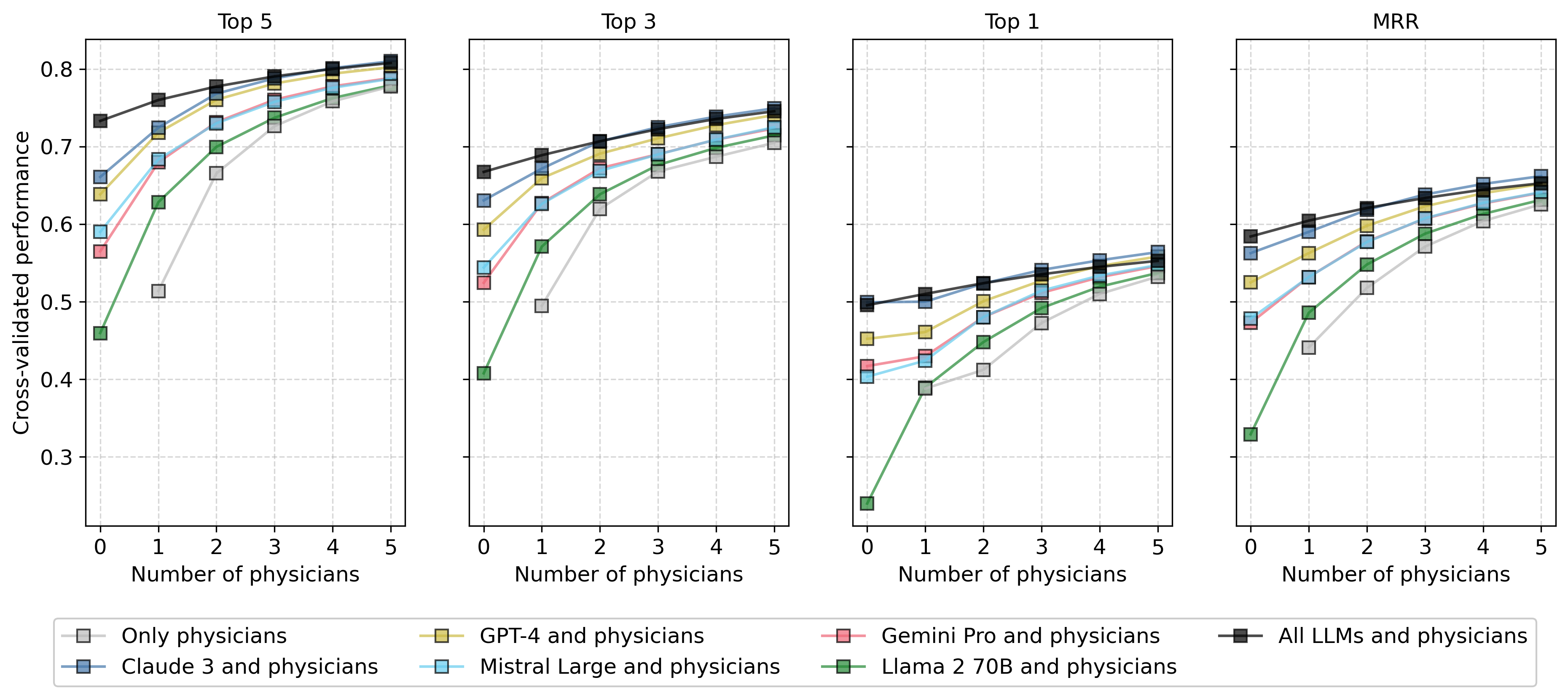}
\caption[Cross-validated performance of human-only ensembles and hybrid ensembles of humans and LLMs]{\textbf{Cross-validated performance of human-only ensembles and hybrid ensembles of humans and LLMs.}
Panels show performance for four outcome metrics ($y$ axes): Top-$k$ indicates the proportion of cases for which the correct diagnosis was among the $k$ top-ranked diagnoses (for $k = \{1,3, 5\}$); MRR shows the mean reciprocal rank of correct diagnoses across cases (see eq. \ref{eq:MRR}).
The individual performance of the five LLMs (and their combined performance in an all-LLMs ensemble) is shown as the left-most square of each color in each panel.
The $x$ axis shows the number of humans added to individual LLMs or to an all-LLMs ensemble.}\label{fig:acc-llm-humans-ensembles}
\end{figure}

Next, we test the complementarity of human and LLM solutions in a hybrid CI approach.
Fig. \ref{fig:acc-llm-humans-ensembles} shows the cross-validated performance when combining the diagnoses of multiple physicians (human-only ensembles as a baseline) with any one of the five individual LLMs or with all LLMs.
For the baseline of human-only ensembles, increasing the number of physicians increased diagnostic accuracy, with greater marginal increases in accuracy for smaller than for larger group sizes. These results are in line with earlier findings from a smaller set of Human Dx cases \cite{kurvers2023, barnett2019comparative}.

Crucially, adding one LLM to the human diagnoses consistently increased performance for both individual physicians and human-only ensembles of different sizes, with the largest increase attained when adding the best-performing individual LLM or an all-LLM ensemble. For top-$5$ and top-$3$ performance metrics, adding the all-LLM ensemble was as good as or better than adding the best-performing LLM. For top-$1$ accuracy and MRR, adding either the best-performing LLM or the all-LLM ensemble yielded the best results---which of the two depended on the size of the human group.  
Even adding the worst-performing LLM, which by itself performed worse than the average individual physician, generally led to a slight increase in performance across all metrics.

From the perspective of human-only ensembles, comparing the performance of ensembles of $n$ humans with that of hybrid ensembles of $n-1$ humans plus one LLM (i.e., the same overall group size of $n$ inputs) showed that adding either the best or second-best LLM or the all-LLM ensemble to a human-only ensemble outperformed adding another human (depending on the accuracy metric and group size, this finding also tended to hold for the third- and fourth-best LLM; Fig.~\ref{fig:acc-llm-humans-ensembles}).
From the perspective of individual LLMs or an all-LLM ensemble, adding one or more human(s) increased performance; this increase was most pronounced for the worst-performing LLMs.

As Extended Data Figs. \ref{fig:hybrid_specialty} and \ref{fig:solvers_humans_students} show, these results held across the five most common medical specialties in our data and for medical students.

\section{Complementarity of human- and LLM-generated diagnoses}

\begin{figure}[!hb]
\centering
\includegraphics[width=1.0\textwidth]{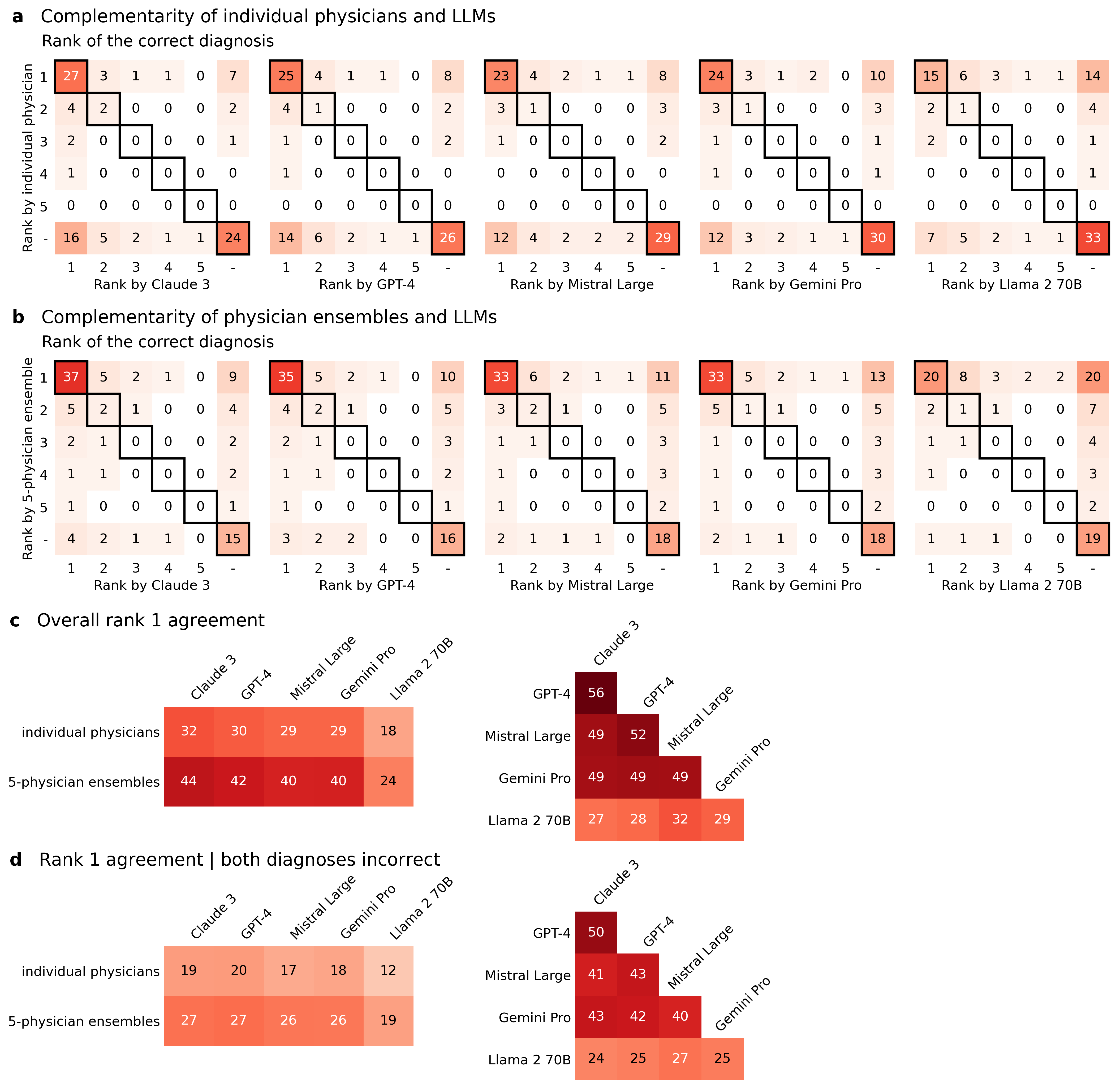}
\caption[Complementarity of diagnoses by individual humans, human-only ensembles and LLMs]{\textbf{Complementarity of solutions from individual humans and human-only ensembles and LLMs.} 
Panels \textbf{a} and \textbf{b} show, for each of the five LLMs, matrices with the percentages of cases for all 36 combinations of the LLM ($x$ axis) and humans ($y$ axis) assigning the correct diagnosis a particular rank (i.e., rank 1, 2, 3, 4, 5 or not ranked). \textbf{a}, Results for individual physicians. \textbf{b}, Results for five-physician human-only ensembles. The highlighted diagonal indicates cases where an LLM and the humans assigned the correct diagnosis the same rank.
Panels \textbf{c} and \textbf{d} show the percentage of cases in which the same diagnoses were assigned rank one, comparing individual physicians and 5-physician ensembles to LLMs (left side), and different LLMs to each other (right side).
\textbf{c}, Overall rank one agreement, regardless of whether the correct diagnosis was included.
\textbf{d}, Rank one agreement when both diagnosticians were incorrect. 
Results were extracted from the cross-validation procedure by recording the frequencies with which physicians and LLMs assigned the same or a different rank to either the correct or an incorrect diagnosis, averaged across all cases and the five folds (see Methods).
Note that due to rounding to integers, there may be small inconsistencies when summing rows or columns across matrices or when comparing sums of values to respective percentages reported in the main text.
}\label{fig:complementarity}
\end{figure}

The results presented in Fig. \ref{fig:acc-llm-humans-ensembles} suggest complementarity of physicians and LLMs in diagnosing open-ended medical problems.
Given that most LLMs outperform the average individual physician, however, how can adding a single physician to an individual LLM---or even to an ensemble of LLMs---increase diagnostic accuracy?
The key answer to this question is that humans and LLMs make different kinds of errors. The literature on both CI \cite{ladha1992condorcet, grofman1983thirteen} and machine ensembles \cite{tumer1996error, kuncheva2014combining} recognizes that the less correlated the errors of its members are, the more successful the ensemble will be.

Fig. \ref{fig:complementarity}a shows the percentage of cases in which individual physicians and LLMs placed the correct diagnosis on the same rank or both did not rank the correct diagnosis (highlighted diagonal cells) and the percentage of cases in which individual physicians and LLMs placed the correct diagnosis on different ranks (or it was only mentioned by either a physician or an LLM; all other cells).
The results show that individual physicians and LLMs did not assign the correct diagnosis to the same rank in a substantial number of cases (range across LLMs:  $46\%$--$51\%$).
Crucially, when LLMs did not list the correct diagnosis at all (range across LLMs: $34\%$--$54\%$;
right-most columns), individual humans did mention it in a substantial number of cases (range across LLMs: $30\%$--$38\%$;
right-most columns excluding bottom-right cells), most frequently ranking it first (range across LLMs: $20\%$--$27\%$, top-right cells). In other words, diagnoses missed by LLMs were often made by individual physicians, frequently in first place.
Thus, although individual physicians performed worse overall than most LLMs (see Fig.~\ref{fig:acc-llm-ensembles} and Extended Data Fig.~\ref{fig:performance_more}), in a substantial number of cases they were able to compensate for the LLMs' errors.
Similarly, when individual humans did not list the correct diagnosis at all ($49\%$, bottom rows),
LLMs did in a substantial number of cases (range across LLMs: $32\%$--$51\%$; bottom columns excluding bottom-right cells), most frequently ranking it first (range across LLMs: $15\%$--$33\%$, bottom-left cells).
%

Fig. \ref{fig:complementarity}b shows the same analysis for five-physician collectives and LLMs. Given that human collectives outperformed individual humans (see Fig.~\ref{fig:acc-llm-humans-ensembles}), the diagnoses given by five-physician collectives are even more complementary to the LLMs than are the ones given by individual physicians.
When LLMs did not list the correct diagnosis at all (range across LLMs: $34\%$--$54\%$; right-most columns), human ensembles did so in the majority of cases (range across LLMs: $55\%$--$65\%$; right-most columns excluding bottom-right cells), most frequently ranking it first (range across LLMs: $26\%$--$36\%$, top-right cells).
%
Intriguingly, the opposite pattern was less pronounced. When human-only ensembles did not list the correct diagnosis at all ($22\%$; bottom rows),
LLMs did so in only the minority of cases (range across LLMs: $17\%$--$32\%$).
%
For a similar complementarity analysis of LLMs with respect to each other, see Extended Data Fig. \ref{fig:complementarity_more}.

Fig. \ref{fig:complementarity}c,d shows how often individual physicians, five-physician ensembles and  LLMs agree with each other on their top-ranked diagnosis.
LLMs agree more among themselves than with physicians (Fig. \ref{fig:complementarity}c) and this difference is particularly pronounced in situations where both human and LLM diagnoses are incorrect (Fig. \ref{fig:complementarity}d).
Furthermore, when humans and LLMs both make errors (Fig. \ref{fig:complementarity}d), they are less likely to assign the same incorrect diagnosis to the first rank compared to their respective overall agreement rate (which includes cases where either or both have ranked the correct diagnosis first; Fig. \ref{fig:complementarity}c).
Extended Data Figs.~\ref{fig:agreements_physicians_LLMs} and \ref{fig:agreements_LLMs_LLMs} show that the above conclusions about error diversity also hold when considering the full range of ranks 1 to 5.
This error diversity is crucial for a CI approach to be effective, and it is significantly more pronounced among hybrid pairs of a physician and an LLM compared to between pairs of different LLMs.
In a collective aggregation scheme based on (weighted) majority voting, this error diversity ensures that correct diagnoses accumulate more frequently than incorrect ones, allowing the correct solutions to rise to the top of the collective differential diagnosis.

\section{Discussion}\label{sec:discussion}

%
Our results demonstrate the potential of combining human medical expertise with large language models (LLMs) to enhance accuracy and reduce errors in open-ended medical diagnostics.
Integrating the differential diagnosis of a single human diagnostician with the output of a single LLM yielded a better performance than either alone. Adding an LLM to multiple physicians' diagnoses also improved performance in nearly all scenarios. The individual accuracy of the LLM influenced the performance gain, with the highest gain from the best-performing LLM. But even the worst-performing LLM, which was less accurate than the average human, showed positive effects.

Taking an LLM perspective, also the performance of LLMs could be boosted by adding human judgements. Adding a single physician increased performance for all LLMs even though individual physicians, on average, performed worse than most LLMs; and LLM performance increased steadily with adding more humans. The increase in performance was highest for the worst-performing LLM and lowest for the best-performing LLM.

An important component of (hybrid human--machine) CI is that different users or machines produce independent and diverse errors \cite{grofman1983thirteen, ladha1992condorcet, marshall2019quorums, steyvers2022bayesian}.
We find that humans and LLMs indeed make complementary errors that disperse throughout the vast solution space, while correct diagnoses accumulate and converge when integrating human and LLM diagnoses. 

Previous work has shown the potential of AI and CI individually, and their hybrid combination for problems with well-defined, small solution spaces (e.g., categorization, probabilistic forecasting, numerical estimation) \cite{Pescetelli2021,Peeters2020,Steyvers2023, Benjamin2023}. Here we showed that these results can be generalized to open-ended problems covering a vast solution space (there are more than 360,000 unique medical concepts in the March 2023 international edition of SNOMED CT that we used), by using a general-purpose method to automatically harmonize and aggregate the solutions generated by humans and LLMs.
While we demonstrated this method in the domain of medical diagnostics, we believe that our approach can be generalized to different applications for which structured domain knowledge is available, allowing the harmonization and principled aggregation of human expert judgements and LLM responses (e.g., climate change adaptation management \cite{Trianni:HHAI2023}).

\section{Limitations and future research}

While our study demonstrates the potential of hybrid human--AI systems in medical diagnostics, further research is necessary to ensure the safety, reliability, efficacy, and ethical deployment of this technology in real-world clinical settings.

For instance, although vignette-based studies represent a validated and accepted paradigm for the study of diagnostic decision-making processes in medicine \cite{peabody2004measuring}, it remains an open question as to how well our method translates to actual clinical practice.
Moreover, our case vignettes were selected by an expert panel at Human Dx, and users may flag suspicious cases for removal from the Human Dx platform. This case selection procedure may have excluded very difficult or rare cases. Future work could consider more ecologically valid or representative ways of selecting cases.

Furthermore, our analyses do not consider the consequences of the treatments implied by the diagnoses. Future work could study whether our proposed approach alters the likelihood of arriving at a potentially beneficial (or harmful) treatment.
Such research must consider the decision context, as the recommended or accessible treatments may vary depending on the cultural, regional, and institutional circumstances, as well as the patients' health insurance plan \cite[e.g.,][]{Yabroff2020}.  

Finally, our study was not designed to address risks related to fairness and equity \cite[see, e.g.,][]{hookerMovingAlgorithmicBias2021,vangiffenOvercomingPitfallsPerils2022,wachterBiasPreservationMachine2021,weidingerEthicalSocialRisks2021,liangHolisticEvaluationLanguage2023,paulus2020predictably}.
For example, LLMs have been shown to perpetuate race-based medicine in their responses \cite{Omiye2023}. This finding suggests that the clinical medical knowledge encoded in LLMs \cite{singhalLargeLanguageModels2023,Jiang2023} is tainted by racism, which can leak into medical diagnoses, resulting in worse health outcomes for disadvantaged groups.
Future work should directly study the extent to which the integration of humans and LLMs mitigates or amplifies the biases of both parties in medical diagnostics (see also \cite{Groh2024}).

More generally, taking a human-centered approach when designing hybrid systems is essential to compensate for the lack of transparency of AI models and for building trust among all affected stakeholders \cite{Birhane2022power, carusiMedicalArtificialIntelligence2023, Delgado2023participatory, wiensNoHarmRoadmap2019}. Such an approach may help identify and mitigate some of the problems of LLMs or hybrid systems already during the design stage.

Future research could build on our approach in several ways.
First, although we used a systematic prompt engineering approach, more sophisticated techniques have been developed that could further boost accuracy (e.g., tree of thought \cite{Yao2023,Long2023}, or self-consistency with temperature/top-$p$ sampling \cite{Wang2022}). Applying sophisticated multi-level prompt-engineering techniques to generalist foundational models can improve performance and even outperform fine-tuned models for the medical domain \cite{nori2023can}. 
Second, more generally, combining computational models of human cognition with LLMs might improve a system's reasoning performance   (see, e.g., \cite{Bhatia2023}, for the case of induction).
Third, vignettes could be classified into categories (e.g., medical specialties, number and type of case findings), and using tailored few-shot examples within these categories when prompting LLMs or adjusting weights for LLMs based on these categories may further boost accuracy.
Additionally, more sophisticated weighting techniques could be tested that adjust weights based on fairness, or LLM biases \cite{paulus2020predictably}.
Fourth, we only considered text-based cases; future work could test the diagnostic performance of large multi-modal models (and hybrid human--AI ensembles) on, for example, images (e.g., x-rays or histopathological images) or sounds (e.g., echocardiograms) alongside the textual information \cite{lu2024multimodal,moorFoundationModelsGeneralist2023}. 
Fifth, future work could further explore the potential of hybrid CI with non-experts. Extended Fig.~\ref{fig:solvers_humans_students} demonstrates that hybrid ensembles of LLMs and medical students were able to outperform individual physicians and even groups of physicians. Boosting the performance of less qualified individuals by leveraging LLMs might have particular potential for underserved regions where access to experts is limited.

\section{Conclusion}

Our study demonstrates the power of hybrid human–AI collectives in the context of medical diagnostics for general clinical practice. Hybrid collectives outperform both individual human experts and LLMs (as well as human-only and LLM-only collectives) in generating accurate differential diagnoses. This superior hybrid performance is a direct consequence of physicians and LLMs making different kinds of errors: when LLMs missed the correct diagnosis, individual physicians often contributed the correct diagnosis, rescuing the hybrid performance. 

Recent years have seen a surge of research and publications on the potential of LLMs (e.g., in medical diagnostics; \cite{singhalLargeLanguageModels2023}). However, in both science and public discourse, there is increasing concern about the lack of safeguards to ensure the safety, quality, and equity of LLM-based systems \cite{chakravortiAITrustProblem2024}. LLMs, despite their impressive capabilities, hallucinate \cite{hong2024hallucinations,Ji2023,Pal2023,tonmoy2024Comprehensive}, lack common sense \cite{williams2024easy}, and are biased \cite{Omiye2023,Navigli2023}---shortcomings that may reflect LLMs' inherent limitations \cite{lenat2023getting} and may thus not be remedied by more sophisticated architectures, more data, or more human feedback. 

We posit that the time has come for a second wave of research on LLMs (and AI in general) that is no longer content to showcase what LLMs can do, propose technical approaches to fix their flaws (e.g., \cite{tonmoy2024Comprehensive}), and speculate about how human oversight could be implemented. Rather, it is crucial to study how to leverage the complementary strengths of humans and AI by combining the experience and common sense of experts with the vast information processed by LLMs. 
In addition to technological solutions aimed at addressing problems inherent in an AI system (e.g., using retrieval-augmented generative AI to try addressing hallucinations; \cite{tonmoy2024Comprehensive}), incorporating complementary human intelligence can help mitigate the risks of LLMs in ways that purely technological solutions may not ever be able to.

\FloatBarrier

\section{Methods}

\subsection{Human Dx: Medical diagnostics cases and data from human solvers}

For our analyses, we used a dataset of 2,133 medical cases with a total of 40,762 diagnoses provided by medical experts through the user interface of the Human Dx app (see Fig. \ref{fig:schema}a). Beforehand, we excluded from our analyses all diagnoses that were incomplete due to submission errors or connectivity issues. We also excluded the diagnoses of users who bypassed the onboarding process and of ``shadow banned'' users, who were permitted access to the platform but excluded from analyses due to unhelpful behavior (e.g., submitting diagnoses consisting of random characters or using profanities). Test accounts belonging to two Human Dx staff members were also excluded.
Finally, we excluded cases containing images (as the LLMs were not able to process these).
The medical experts consisted of 1,370 attending physicians ($37.3\%$), 139 fellows ($3.8\%$), and 2,160 resident physicians ($58.9\%$), representing senior doctors, doctors undergoing specialized training, and doctors in training, respectively. Note that this tenure information is based on self-reports by the users. As Extended Data Fig.~\ref{fig:solvers_humans_expertise}a shows, the performance distributions of these three tenure levels were similar; we therefore combined them into a common category labelled ``physicians.''
An additional 11,772 diagnoses were contributed by 1,037 medical students; on average, these were less accurate (Extended Data Fig.~\ref{fig:solvers_humans_expertise}a,b). For hybrid human--LLM ensembles (see Fig. \ref{fig:acc-llm-humans-ensembles}), only cases diagnosed by a minimum of five physicians were analyzed (so that collectives of up to five humans could be simulated), totaling 1,928 cases.
The medical specialty of a case (used for the robustness analyses reported in Extended Data Figs. \ref{fig:LLMs_specialty} and \ref{fig:hybrid_specialty}) was determined by prompting \llmClaude\ to identify the three most probable specialties from a list of 145 specialties used internally by Human Dx (see SI for the exact wording of the prompt). Only the most probable specialty was used in the analyses shown in Extended Data Figs. \ref{fig:LLMs_specialty} and \ref{fig:hybrid_specialty}.

\subsection{LLMs: Prompt engineering and post-processing of responses}\label{sec:prompt_engineering}

Prompt engineering can markedly affect the quality and format of LLM responses. There is no established framework for prompt engineering, and which wording produces the desired response typically depends on the LLM used. 
Some studies have found that shorter prompts work better \cite{zhang2022automatic}; others that complex prompts yield better responses \cite{fu2022complexity}. In practice, prompts are generally engineered by trial and error \cite{fu2022complexity}.

We took a systematic, semi-exhaustive approach, building up prompts in a modular fashion by concatenating several text blocks (Extended Data Fig. \ref{fig:prompt_engineering}). The most basic block feeds the case vignette to the LLM verbatim. The case vignette describes the patient's symptoms, test results, and medical record. The LLM is then asked to provide the five most probable diagnoses ordered by their likelihood of being correct (i.e., a differential diagnosis). We included several additional text blocks in the prompt and tested whether these additions increased diagnostic accuracy. The prompt that performed best in a training fold of cases was then used for the analysis in the remaining folds.
Specifically, the additional text blocks assign
the LLM the role of a medical expert (impersonation \cite{Salewski2023}), advise it to check that the proposed diagnoses are consistent with the case description (self-consistency), advise the LLM to report diagnoses in SNOMED CT terminology (answer format SCT) or in common shorthand (answer format common), or offer five examples of case vignettes with their correct diagnoses (a technique known as few-shot prompting; \cite{brown2020language}). In selecting the few-shot examples, we sought to ensure variety in patients' age (5 months to 89 years) and gender (3 female, 2 male) and the medical specialty.
The resulting LLM responses constitute the basis of the results reported here.
For details of the exact wording of prompts and results of the validation process, see Extended Data Fig. \ref{fig:prompt_engineering} and SI.

Our general validation approach is as follows: We used five-fold cross-validation on the whole set of cases, using one fold of cases to select the best prompt and calculate the weights for humans and LLMs (see ``\nameref{sec:weighting}''). The other four folds were used for assessing out-of-sample performance. We report results averaged across the five cross-validation outcomes. 

The raw LLM responses required some additional post-processing (which was not needed for the human responses). Even when explicitly instructed to provide answers in a specific format, some LLMs did not always comply and occasionally returned verbose responses. However, these responses follow typical patterns that are easy to recognize. Some LLMs, for example, start the response with an introductory sentence before parsing the differential diagnosis in the requested format. We therefore removed the response until the first line break if the response started with ``Sure,...,'' ``Here is the...,'' ``Here are...,'' ``\#\#\# Response:...,'' ``The probable...,'' ``The differential...,'' ``The most probable...,'' or ``Based on...''. Furthermore, we removed various forms of list numbering.

\subsection{Matching raw text to unique medical concepts (SNOMED CT)}
One of the main challenges when aggregating individual diagnoses in open-ended medical diagnostics is discerning which diagnoses correspond to the same medical concept. 
The differential diagnoses given by humans and LLMs consist of raw text. Two strings pointing to the same disease might differ slightly---for example, due to typos, use of synonyms, or differences in spelling.
To facilitate comparison of these open-ended diagnoses, we developed a method and processing pipeline that leveraged the comprehensive SNOMED CT healthcare terminology (March 2023 International Edition Release) and mapped the raw string responses to unique IDs in SNOMED CT (extending a pipeline described in \cite{kurvers2023}).

The first step is string normalization, using routine natural language processing  tools to standardize all diagnoses---including the correct ones provided by cases' authors. The normalization procedure involves removing stop words, converting British English to US English, converting plural to singular, and identifying acronyms; specifically, we used the \textit{Norm}\footnote{https://lhncbc.nlm.nih.gov/LSG/Projects/lvg/current/docs/userDoc/tools/norm.html} pipeline, one of the \textit{Lexical Tools} maintained by the \textit{National Library of Medicine}.
The second step is to map concepts to SNOMED CT IDs (Fig. \ref{fig:schema}c). This is done by comparing a normalized diagnosis string to the normalized entries in SNOMED CT including all of their stored synonyms sharing the same ID. A SNOMED CT ID is assigned to a diagnosis only when there is an exact match between the sets of words---in other words, the compared strings having a Jaccard similarity of $1$.
On the rare occasion that more than one SNOMED CT ID is matched by this technique, SNOMED CT allows for differentiation by semantic tags. We gave preference to SNOMED IDs according to their semantic tags in the following order: ``disorder,'' ``finding,'' ``morphologic abnormality,'' ``body structure,'' ``person,'' ``organism,'' ``specimen'' (see \cite{kurvers2023} for the rationale behind this ordering), so that a diagnosis was only matched to exactly one ID. 

Applying this approach, as described in \cite{kurvers2023}, produced a match for $90\%$ of the correct case diagnoses, $78\%$ of diagnoses given by LLMs (calculated across all prompts), and $84\%$ of diagnoses given by humans. 
For the diagnoses that could not be matched, we employed a different approach. We created 768-dimensional vector embeddings of all unique (active) SNOMED CT concepts and synonyms using a sentence-transformer model based on the \textit{pubmedbert} model \cite{pubmedbert}---a domain-specific transformer model trained on texts from the \textit{National Library of Medicine}  and  fine-tuned over the \textit{MS-MARCO} dataset using the sentence-transformer framework \cite{deka2022improved}.
We then created a vector embedding of the diagnosis to be matched and assigned it the SNOMED CT ID for which the cosine similarity between embedding vectors was highest. We were thus able to match all remaining raw string diagnoses to exactly one SNOMED CT ID. For example, the diagnosis ``Chlamydia infection'' which could not be matched before was now correctly matched to the SNOMED CT concept ``Chlamydial infection (disorder).'' Likewise, ``Human immunodeficiency virus disease'' was correctly matched to the SNOMED CT concept ``Human immunodeficiency virus infection (disorder).'' As a sanity check, we applied the sentence-transformer matching technique to all diagnoses that were successfully matched in the first approach (i.e., using the pipeline described in \cite{kurvers2023}) and found that both methods arrived at the same SNOMED CT ID for $99.4 \%$ of diagnosis strings (given by humans or LLMs).

\subsection{Weighted aggregation of LLMs and/or human inputs}\label{sec:weighting}

To aggregate individual diagnoses into a collective diagnosis, we implemented a scoring rule. After normalizing all differential diagnoses and matching them to unique SNOMED CT IDs, we built a set of all nominated IDs (see Fig. \ref{fig:schema}b--d and previous subsection). Then, for each diagnostician (physician or LLM) and each diagnosis, a partial score was assigned that was discounted depending on the rank $r$ in the differential diagnosis (i.e., the list of diagnoses ordered in descending order of judged probability of being the correct diagnosis).
Following \cite{barnett2019comparative, kurvers2023}, we employed a $1/r$ rule for the rank-discounted partial score (i.e., the inverse rank of a diagnosis). 
Additionally, this partial score was multiplied by a weight at the level of the diagnostician (see next paragraph). Finally, for each nominated diagnosis, these partial scores were summed up over all diagnosticians, and the ranking of the collective differential diagnosis was defined as a list sorted in decreasing order of the overall score a diagnosis received.

Prior research on CI in medical diagnostics has shown that giving equal weight to members in a collective when aggregating individual judgements into a collective diagnosis (i.e., using a simple equal-weighting combination rule) performs well as long as there is not much difference in individual performance \cite{kurvers2016boosting}. However, if there are substantial differences in individual accuracy, giving the more competent individuals higher weights in the aggregation step may improve performance. 
We therefore used the Weighted Majority Voting Ensemble (WMVE) approach described in \cite{Dogan2019} to determine weights for LLMs and humans. 
Weights were determined on one-fifth of the cases and calculated for each configuration (i.e., combinations for the accuracy metric used and which LLMs and/or the number of human experts). The performance of the WMVE was then calculated on the remaining four-fifths of the cases. Results are reported as the means of a five-fold cross-validation (see also ``\nameref{sec:prompt_engineering}''). 
At the start of the weight-learning process, each diagnostician $j$ in an ensemble of $n$ diagnosticians (physicians or LLMs) is assigned a weight of $w_{j,0} = 1$. For each case $i$ in the training set, the weights are updated according to $w_{j,i} = w_{j,i-1} + \alpha_i$, where $\alpha_i = s_{j,i} \cdot (n - \sum_{j=1}^n s_{j,i})/n$ and $s_{j,i}$ is the score of diagnostician $j$ achieved on case $i$, which depends on the performance metric used (for top-$k$, it is either 1 or 0; for reciprocal rank, it is $1/r$); that is, we estimated weights separately for each metric we evaluated. This means that the weight increases if a diagnostician correctly diagnoses a case in the training set, with a larger increase if the diagnoses of other diagnosticians in the ensemble are incorrect. 
It was not possible to calculate a weight for each individual physician because many only rated a few (or none) of the cases in the training set. We therefore calculated a shared, average weight for all physicians. To this end, for each case in the training set and for each hybrid configuration with $n$ humans, we built all possible groups of $n$ physicians (i.e., using the physicians who provided a differential diagnosis for that case) and averaged over them. If the number of possible groups exceeded 100, we randomly sampled 100 unique groups. 
In most cases, applying such a weighted combination rule outperformed a simple equal-weighting combination rule. However, even with equal weights applied, LLM and hybrid ensembles generally outperformed individual LLMs and physicians (see Extended Data Fig.~\ref{fig:llm-accuracy-weighted-vs-unweighted}).

\backmatter

\begin{appendices}

\section{Data availability}
We include one Human Dx case along with the differential diagnoses provided by humans and LLMs to illustrate our approach which can be found at \url{https://github.com/nikozoe/human_ai_collectives}. The full dataset cannot be shared publicly due to privacy and data protection regulations, but can be obtained by contacting Human Dx.

\section{Code availability}
The code used to run simulations and perform analyses is available at: \url{https://github.com/nikozoe/human_ai_collectives}.

\section{Acknowledgements}
We thank the Human Dx team for providing the data and supporting this research. This work was funded by the Max Planck Institute for Human Development, the European Union’s Horizon Europe research and innovation programme within the context of the project HACID (GA 101070588), and the Deutsche Forschungsgemeinschaft (DFG, German Research Foundation) under Germany's Excellence Strategy–EXC 2002/1 ``Science of Intelligence''–project number 390523135.
We thank Susannah Goss for editing the manuscript.

\section{Author information}

These authors jointly supervised this work: Ralf H.J.M. Kurvers, Stefan M. Herzog.

\subsection{Authors and Affiliations}
\noindent \textbf{\orgdiv{Center for Adaptive Rationality}, \orgname{Max Planck Institute for Human Development}, \orgaddress{\state{Berlin}, \country{Germany}}} \\
\fnm{Nikolas} \sur{Z{\"o}ller}, \fnm{Julian} \sur{Berger}, \fnm{Ralf H.J.M.} \sur{Kurvers}, \fnm{Stefan M.} \sur{Herzog}

\noindent\textbf{\orgname{The Human Diagnosis Project}, \orgaddress{\city{San Francisco} , \state{CA}, \country{USA}}} \\
\fnm{Irving} \sur{Lin}, \fnm{Nathan} \sur{Fu}, \fnm{Jayanth} \sur{Komarneni}, \fnm{Kyle} \sur{Laskowski}

\noindent\textbf{\orgname{University of Cologne}, \orgaddress{\city{Cologne},\country{Germany}}} \\
\fnm{Gioele} \sur{Barabucci}

\noindent\textbf{\orgname{Harvey Mudd College}, \orgaddress{\city{Claremont}, \state{CA}, \country{USA}}} \\
\fnm{Victor} \sur{Shia}

\noindent\textbf{\orgname{Oxford University}, \orgaddress{\city{Oxford}, \country{UK}}} \\
\fnm{Benjamin} \sur{Harack}

\noindent\textbf{\orgname{Kaiser Permanente}, \orgaddress{\city{Downey}, \state{CA}, \country{USA}}} \\
{\fnm{Eugene A.} \sur{Chu}}

\noindent\textbf{\orgname{ISTC-CNR}, \orgaddress{\city{Roma}, \country{Italy}}}\\
\fnm{Vito} \sur{Trianni}

\noindent\textbf{\orgdiv{Science of Intelligence Excellence Cluster}, \orgname{ Technical University Berlin}, \orgaddress{ \city{Berlin}, \country{Germany}}} \\
\fnm{Ralf H.J.M.} \sur{Kurvers}


\subsection{Contributions}

Following the CRediT standard \cite{holcombe2020documenting}.

\begin{itemize}

\item \emph{Conceptualization}: N.Z., J.B., I.L., N.F., J.K., G.B., V.S., B.H., V.T., R.H.J.M.K., and S.M.H. 

\item \emph{Data curation}: N.Z., J.B., I.L., N.F., and G.B.  

\item \emph{Formal analysis}: N.Z., J.B., I.L., N.F., G.B., and S.M.H.  

\item \emph{Funding acquisition}: I.L., J.K., G.B., V.T., R.H.J.M.K., and S.M.H.  

\item \emph{Investigation}: N.Z., J.B., I.L., N.F., and G.B.  

\item \emph{Methodology}: N.Z., J.B., I.L., N.F., G.B., K.A.L., V.S., B.H., E.A.C., V.T., R.H.J.M.K., and S.M.H.  

\item \emph{Project administration}: N.Z., I.L., N.F., J.K., G.B., V.T., R.H.J.M.K., and S.M.H.  

\item \emph{Resources}: I.L., N.F., and K.A.L.  

\item \emph{Software}: N.Z., J.B., I.L., N.F., G.B., K.A.L., V.S., and B.H.  

\item \emph{Supervision}: I.L., J.K., G.B., E.A.C., V.T., R.H.J.M.K., and S.M.H.  

\item \emph{Validation}: N.Z., J.B., N.F., G.B., V.T., R.H.J.M.K., and S.M.H.  

\item \emph{Visualization}: N.Z., J.B., G.B., and S.M.H.  

\item \emph{Writing - original draft}: N.Z., J.B., G.B., R.H.J.M.K., and S.M.H.  

\item \emph{Writing - review \& editing}: N.Z., J.B., I.L., N.F., J.K., G.B., K.A.L., V.S., B.H., E.A.C., V.T., R.H.J.M.K., and S.M.H.

\end{itemize}

\subsection{Corresponding authors}
Correspondence to Nikolas Z{\"o}ller, Ralf H.J.M. Kurvers or Stefan M. Herzog.

\section{Ethics declarations}
We did not collect data explicitly for this study; instead, we analyzed existing data provided by Human Dx. When users sign up on the Human Dx platform, they give consent for their data to be processed and analyzed for research purposes. We consulted the Ethics Committee of the Max-Planck Institute for Human Development and one of their representatives provided guidelines for study procedures. We have complied with all relevant ethical regulations regarding data protection.

\subsection{Competing interests}
Irving Lin and Jayanth Komarneni have personal financial interests in Human Dx. Nathan Fu, Gioele Barabucci and Kyle A. Laskowski are Human Dx consultants.
Victor Shia, Benjamin Harack and Eugene A. Chu were previously employed by Human Dx.


\pagebreak
\section{Extended data figures and tables}
\setcounter{figure}{0}  
\counterwithin{figure}{section}  
\renewcommand{\thefigure}{S\arabic{figure}} 
\captionsetup[figure]{name=Extended Data Fig.}
\setcounter{table}{0}
\counterwithin{table}{section}
\renewcommand{\thetable}{S\arabic{table}} 
\captionsetup[table]{name=Extended Data Table}
\begin{figure}[th]
\centering
\includegraphics[width=0.65\textwidth]{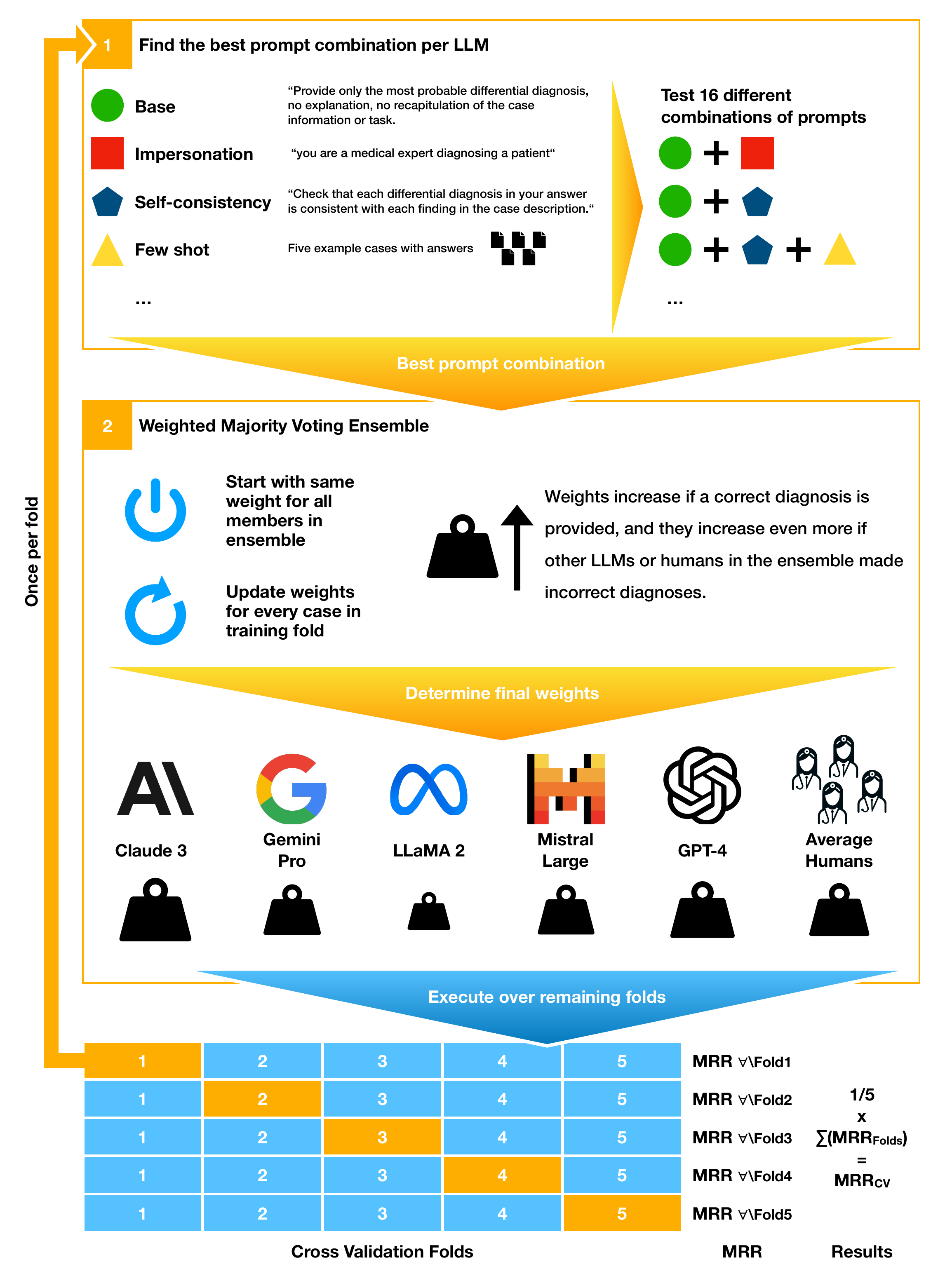}
\caption[Illustration of LLM prompt engineering and validation method]{\textbf{Illustration of LLM prompt engineering and validation method.} We nested our prompt engineering and Weighted Majority Voting Ensemble (WMVE) \cite{Dogan2019} sequence in a five-fold cross-validation procedure. First, we determined which prompt performed best for each LLM in the training fold (one-fifth of the data; see Methods). Second, we calculated weights for each member of the ensemble, also using the training fold. The weights were then used to aggregate collective diagnoses and evaluate the ensemble's performance on the remaining folds (four-fifths of the data). This process yields one result per fold, of which the averages are reported in the main text. We repeated this procedure for every metric reported in the main text (i.e., top-1, top-3, top-5 and MRR).
}\label{fig:prompt_engineering}
\end{figure}
\FloatBarrier
\pagebreak

\begin{figure}[th]
\centering
\includegraphics[width=.9\textwidth]{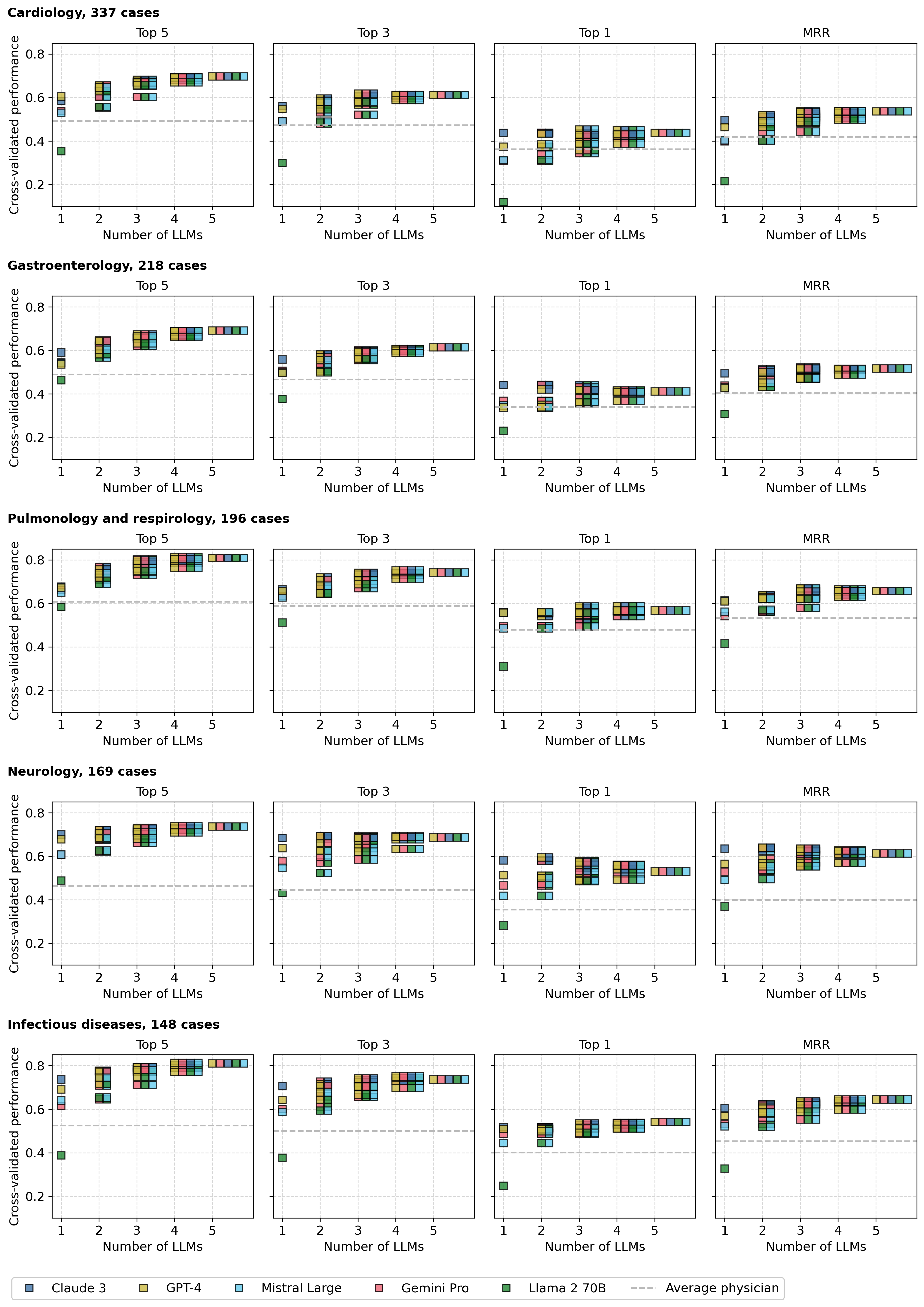}
\caption[Cross-validated performance of LLMs and LLM ensembles by specialty]{\textbf{Cross-validated performance of five individual LLMs and ensembles of all possible combinations of LLMs for the five most common specialties in the dataset}. Across all medical specialties, combining several LLMs into a collective increased diagnostic accuracy relative to the best-performing individual LLM across all performance metrics except top-$1$. In most cases, the best results were obtained by combining all LLMs. \label{fig:LLMs_specialty}}
\end{figure}
\pagebreak

\begin{figure}[th]
\centering
\includegraphics[width=.75\textwidth]{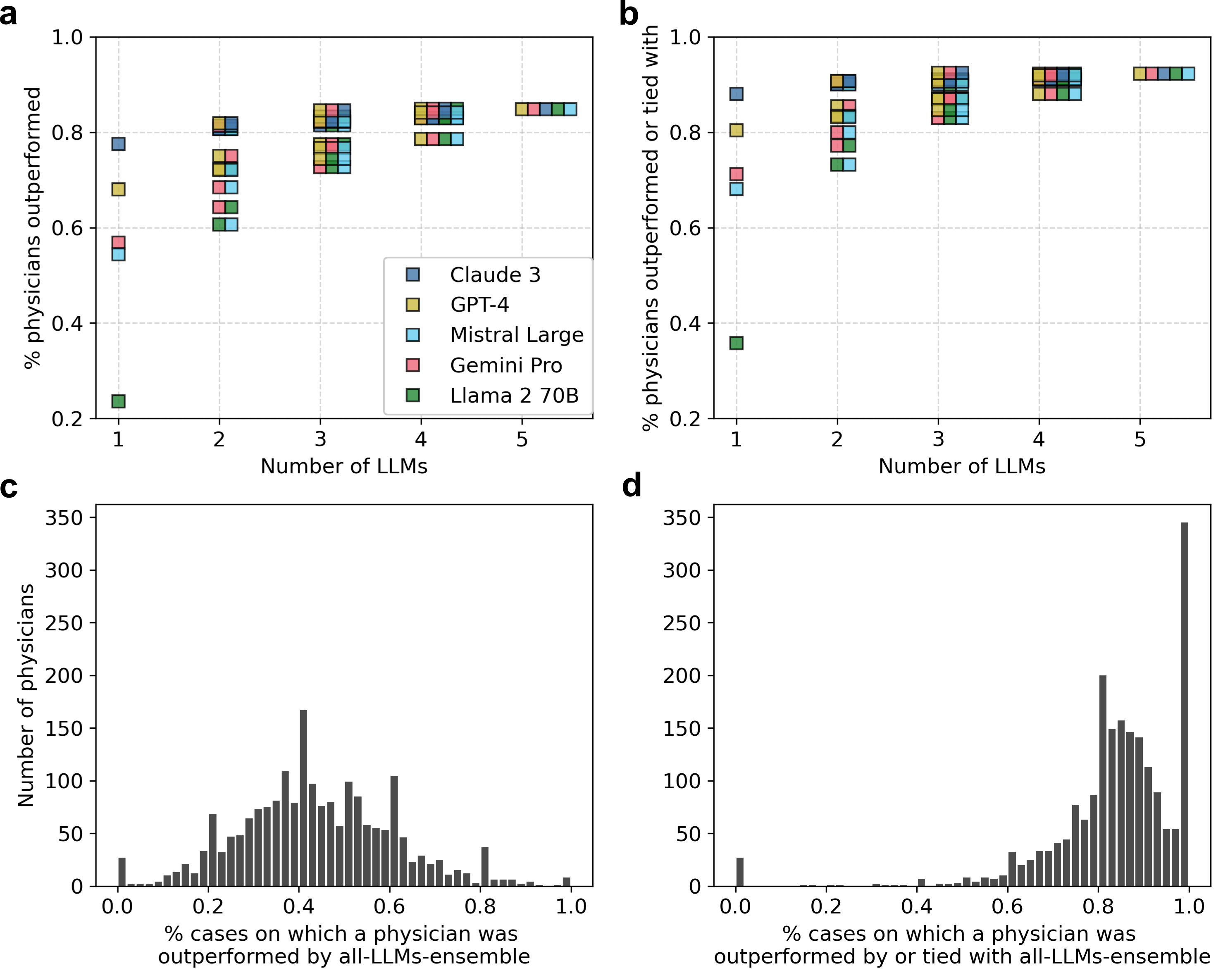}
\caption[Cross-validated relative performance of individual humans and LLMs]{\textbf{Cross-validated relative performance of individual humans and LLMs.} \textbf{a}, Percentage of physicians outperformed by an LLM across the cases they solved. The analysis was limited to physicians who diagnosed five or more cases (n = 1,997). A physician was outperformed on a case if the LLM (ensemble) placed the correct diagnosis at a higher rank; a physician was counted as outperformed overall if they were outperformed more often than they outperformed the LLM (ensemble) across their set of solved cases. \textbf{b}, Percentage of physicians outperformed by or tied with an LLM (ensemble) across the cases they solved.
\textbf{c} and \textbf{d}, Results for the LLM ensemble only. \textbf{c}, Histogram of the number of physicians outperformed on a certain percentage of cases by the LLM ensemble. \textbf{d}, Histogram of the number of physicians outperformed or tied with on a certain percentage of cases by the LLM ensemble.
\textbf{c} and \textbf{d} differ considerably due to the significant number of ties on the case level. 
}\label{fig:performance_more}
\end{figure}
\pagebreak

\begin{figure}[th]
\centering
\includegraphics[width=.88\textwidth]{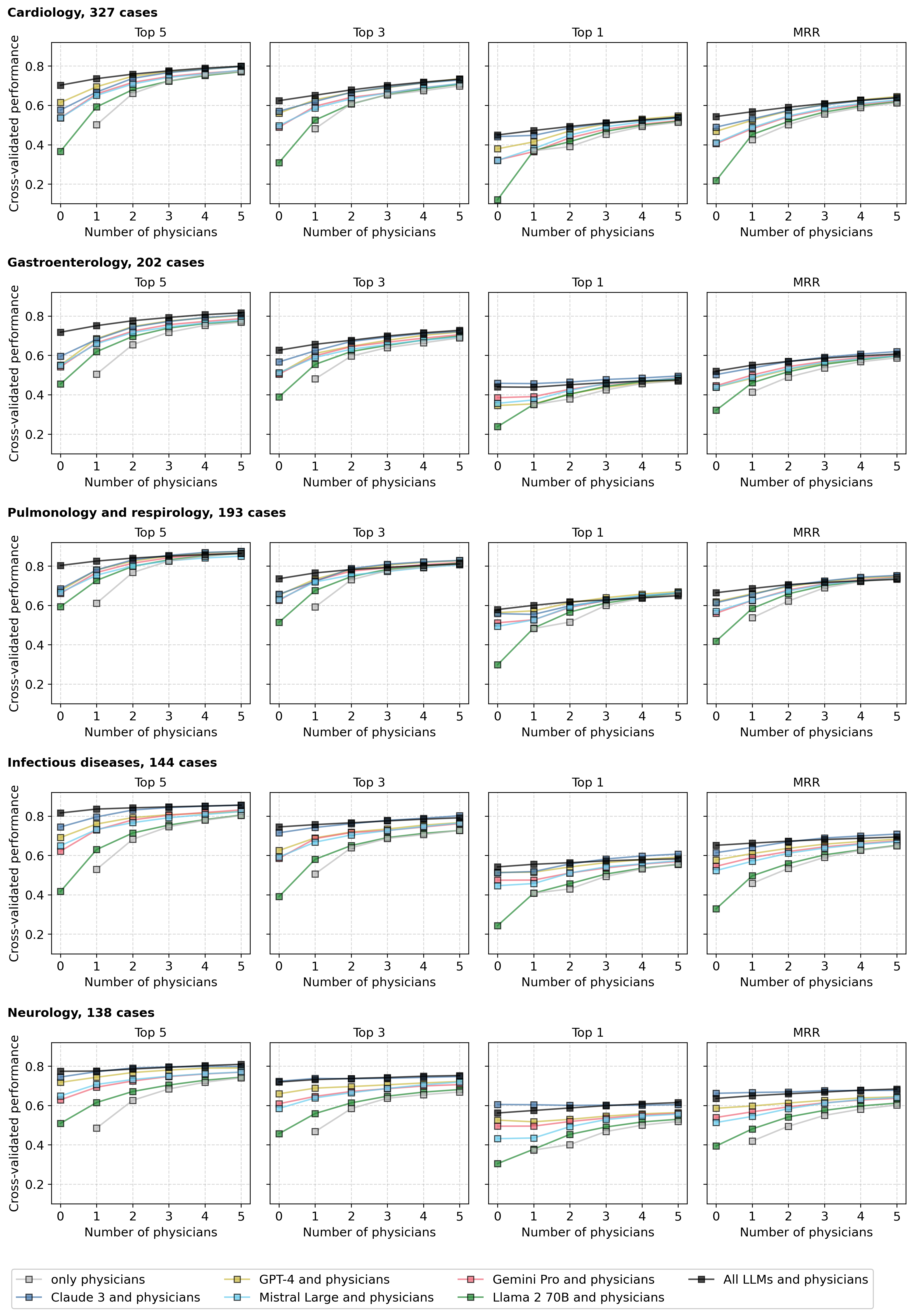}
\caption[Cross-validated performance of human-only ensembles and hybrid ensembles of humans and LLMs by specialty]{\textbf{Cross-validated performance of human-only ensembles and hybrid ensembles of humans and LLMs for the five most common specialties in the dataset}.
Across all medical specialties, combining humans and LLMs increased diagnostic accuracy relative to individual humans or LLMs. Increasing the number of humans in the ensemble generally increased performance. The best top-$5$ and top-$3$ accuracies were achieved by adding all LLMs to the ensemble, the best top-$1$ or MRR performance was achieved by adding either all LLMs or, in some cases, the best-performing individual LLM.
}\label{fig:hybrid_specialty}
\end{figure}
\pagebreak

\begin{figure}[th]
\centering
\includegraphics[width=1.0\textwidth]{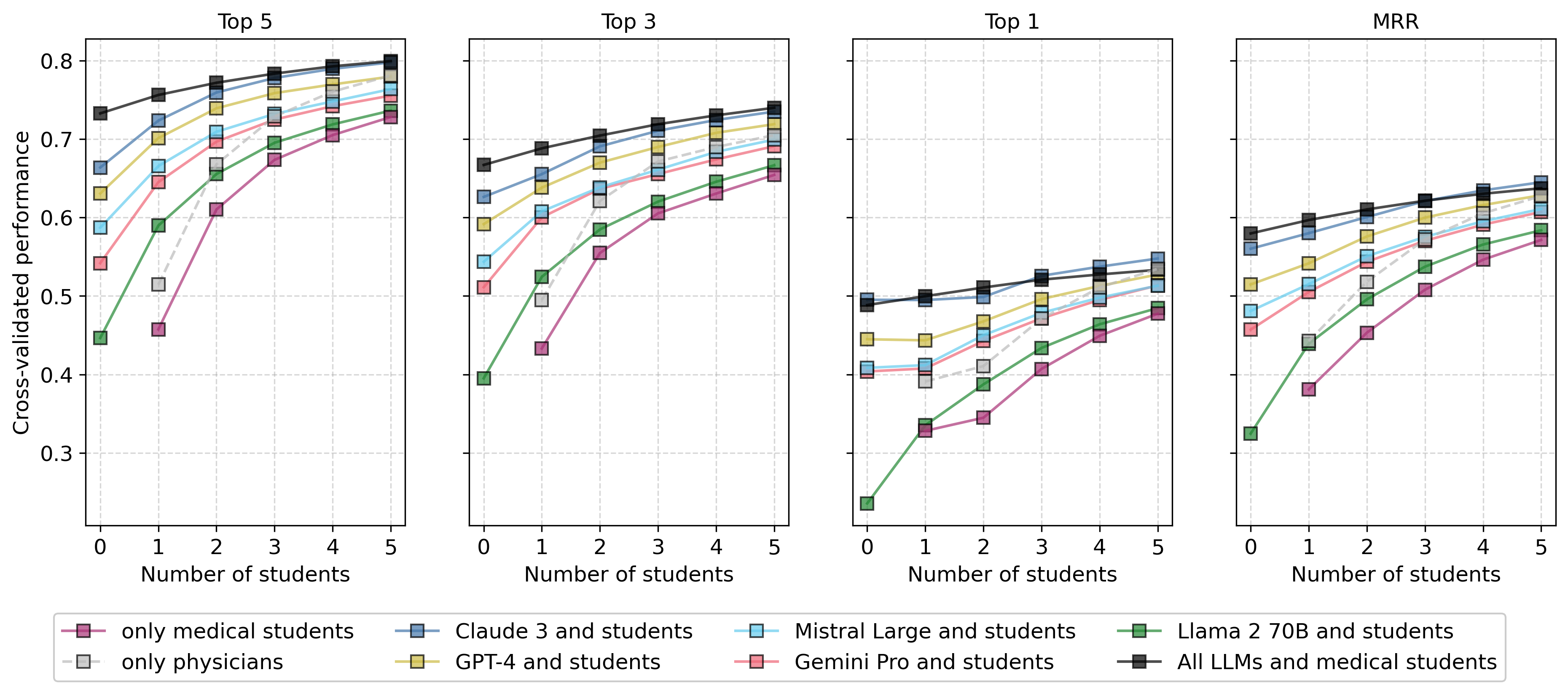}
\caption[Cross-validated performance of medical-student-only ensembles and hybrid ensembles of medical students and LLMs]{
\textbf{Cross-validated performance of medical-student-only ensembles and hybrid ensembles of medical students and LLMs.}
Panels show performance for four outcome metrics ($y$ axes): Top-$k$ indicates the proportion of cases for which the correct diagnosis was among the $k$ top-ranked diagnoses (for $k = \{1,3, 5\}$); MRR shows the mean reciprocal rank of correct diagnoses across cases (see eq. \ref{eq:MRR}).
The individual performance of the five LLMs (and their combined performance in an all-LLMs ensemble) is shown as the left-most square of each color in each panel.
The $x$ axis shows the number of medical students added to individual LLMs or an all-LLMs ensemble. For comparison, the gray boxes and lines show the performance of physicians on the same set of cases. Results are based on 974 medical cases, each diagnosed by at least five medical students. 
}\label{fig:solvers_humans_students}
\end{figure}
\pagebreak

\begin{figure}[th]
\centering
\includegraphics[width=1\textwidth]{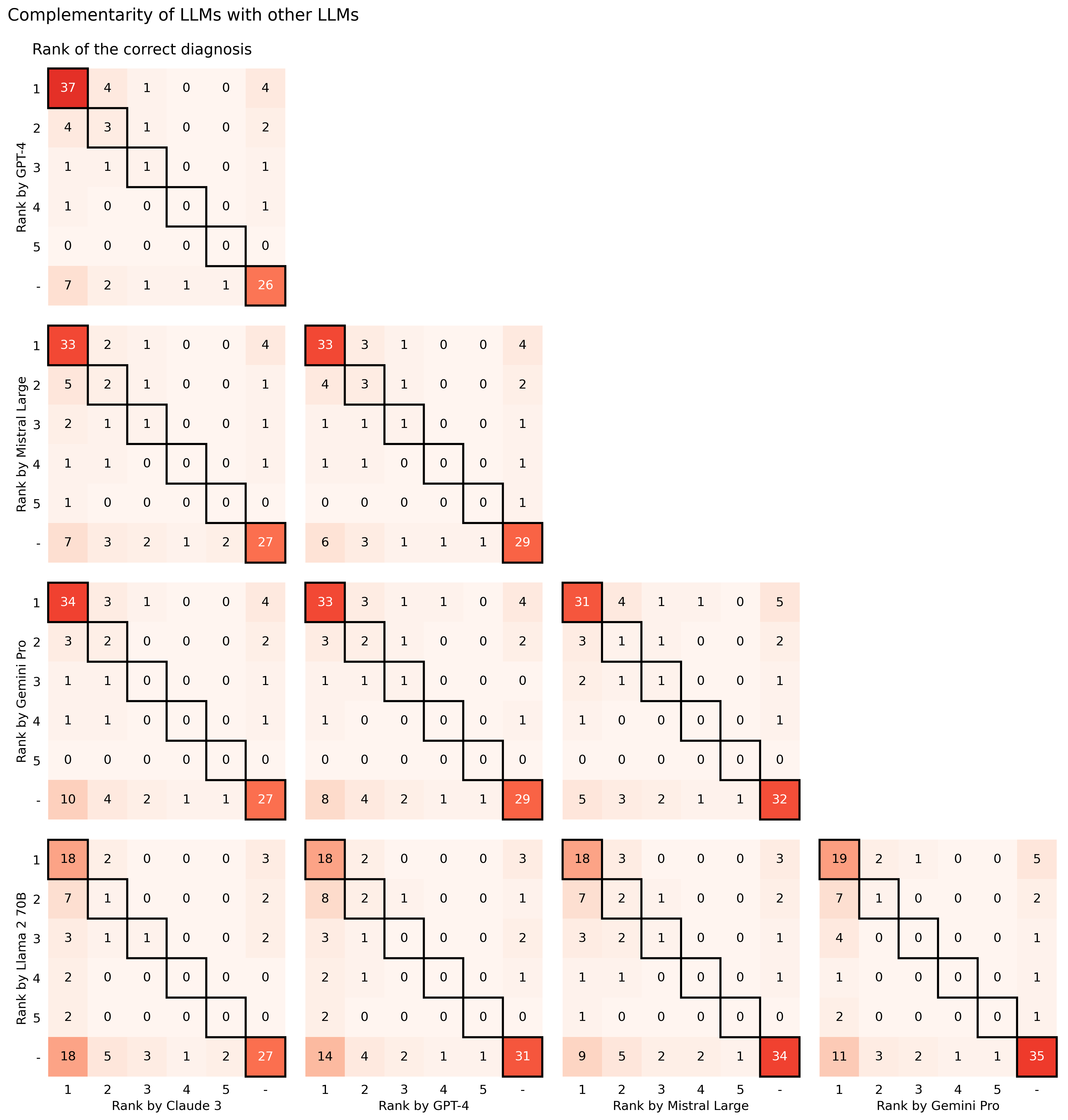}
\caption[Complementarity of solutions among LLMs: Rank of correct diagnoses within the differential diagnoses of LLMs]{\textbf{Complementarity of solutions among LLMs: Rank of correct diagnoses within the differential diagnoses of LLMs.} 
Panels show, for each of the ten possible pairs of LLMs, matrices with the percentages of cases for all 36 combinations of the
LLM at the top ($x$ axis) and LLMs on the side ($y$ axis) assigning the correct diagnosis a particular rank (i.e., rank 1, 2, 3, 4, 5, or not ranked). The highlighted diagonal shows cases where LLMs assigned the same rank to the correct diagnosis. Results were extracted from the cross-validation procedure by recording the frequencies with which
the LLMs assigned the correct diagnosis the same or a different rank, averaged across all cases and the five folds (see Methods).  Note that due to rounding to integers, there may be small inconsistencies when summing rows or columns across matrices.
}\label{fig:complementarity_more}
\end{figure}
\pagebreak

\begin{figure}[th]
\centering
\includegraphics[width=1\textwidth]{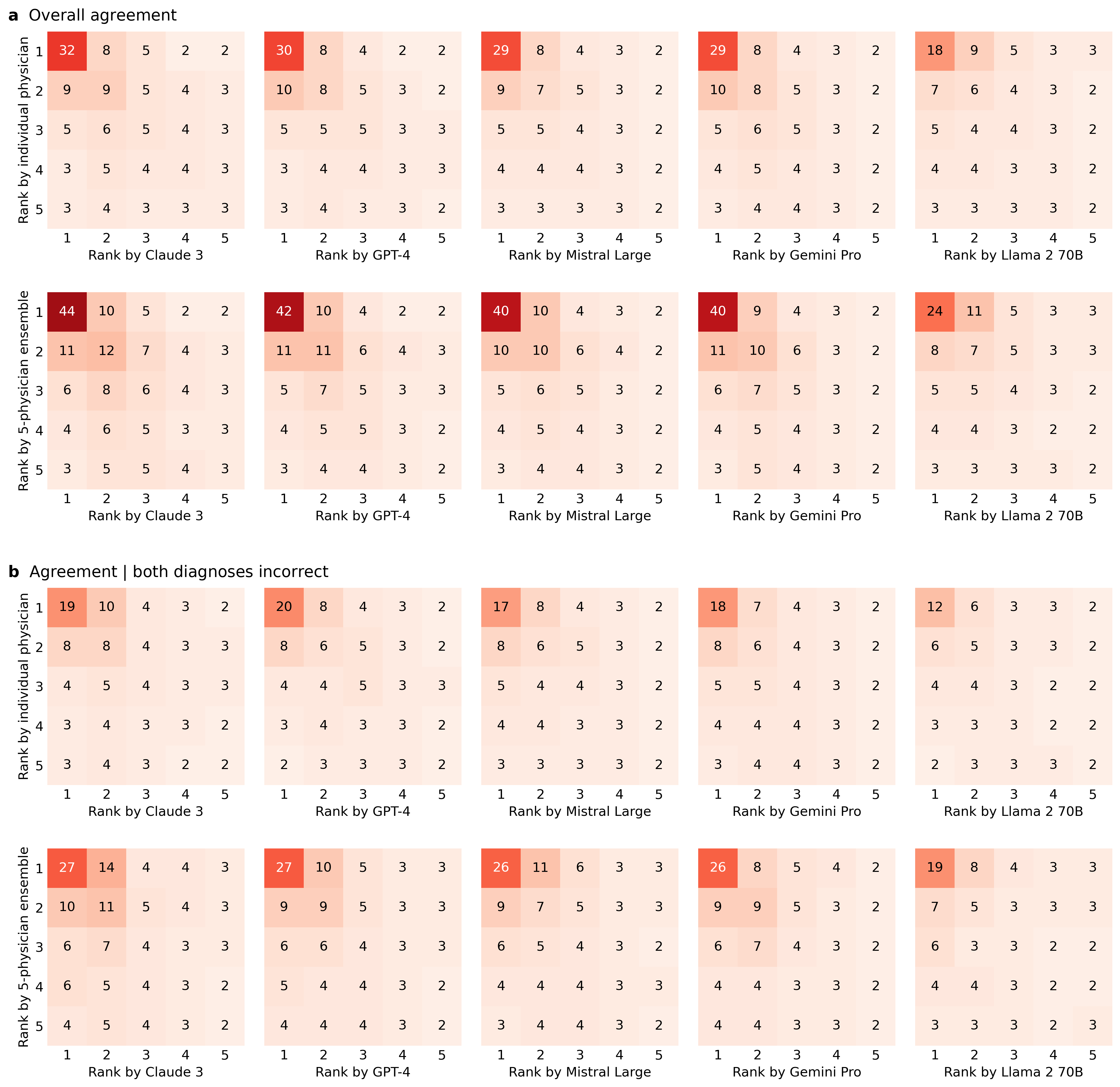}
\caption[Agreement among physicians and LLMs]{\textbf{Agreement among physicians and LLMs.} 
Panels show the percentage of cases in which the same diagnoses were assigned to a particular rank combination, comparing individual physicians and 5-physician ensembles to LLMs.
\textbf{a}, Overall agreement, regardless of whether the correct diagnosis was included in a particular rank combination.
\textbf{b}, Agreement when both diagnosticians were incorrect for a rank combination. 
Results were extracted from the cross-validation procedure by recording the frequencies with which physicians and LLMs assigned the same diagnosis to a rank combination, averaged across all cases and the five folds (see Methods).
}\label{fig:agreements_physicians_LLMs}
\end{figure}

\begin{figure}[th]
\centering
\includegraphics[width=1\textwidth]{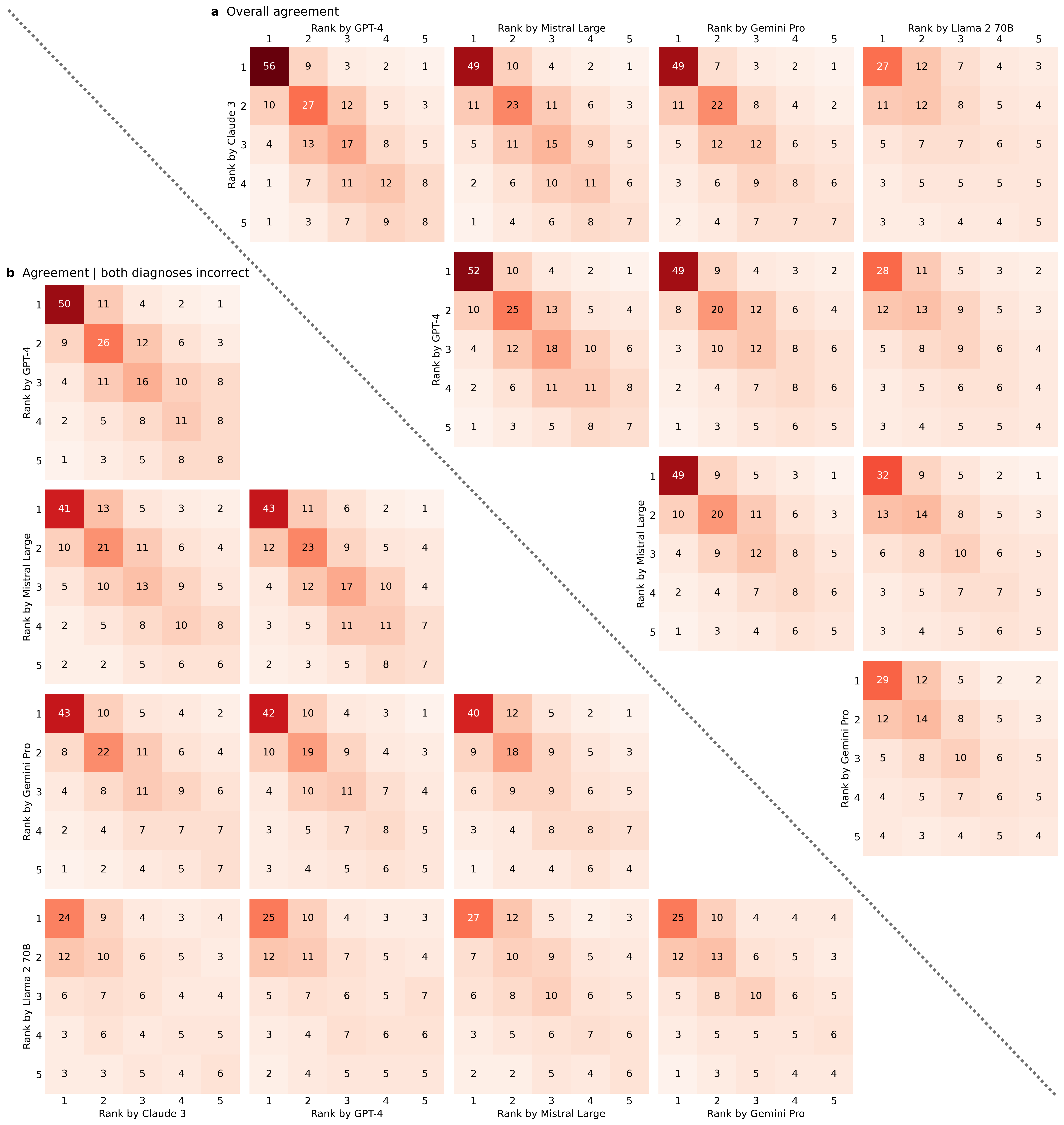}
\caption[Agreement among LLMs]{\textbf{Agreement among LLMs} 
Panels show the percentage of cases in which the same diagnoses were assigned to a particular rank combination, comparing different LLMs to each other.
\textbf{a}, Overall agreement, regardless of whether the correct diagnosis was included in a particular rank combination.
\textbf{b}, Agreement when both LLMs were incorrect for a rank combination. 
Results were extracted from the cross-validation procedure by recording the frequencies with LLMs assigned the same diagnosis to a rank combination, averaged across all cases and the five folds (see Methods).
}\label{fig:agreements_LLMs_LLMs}
\end{figure}
\pagebreak

\begin{figure}[th]
\centering
\includegraphics[width=1\textwidth]{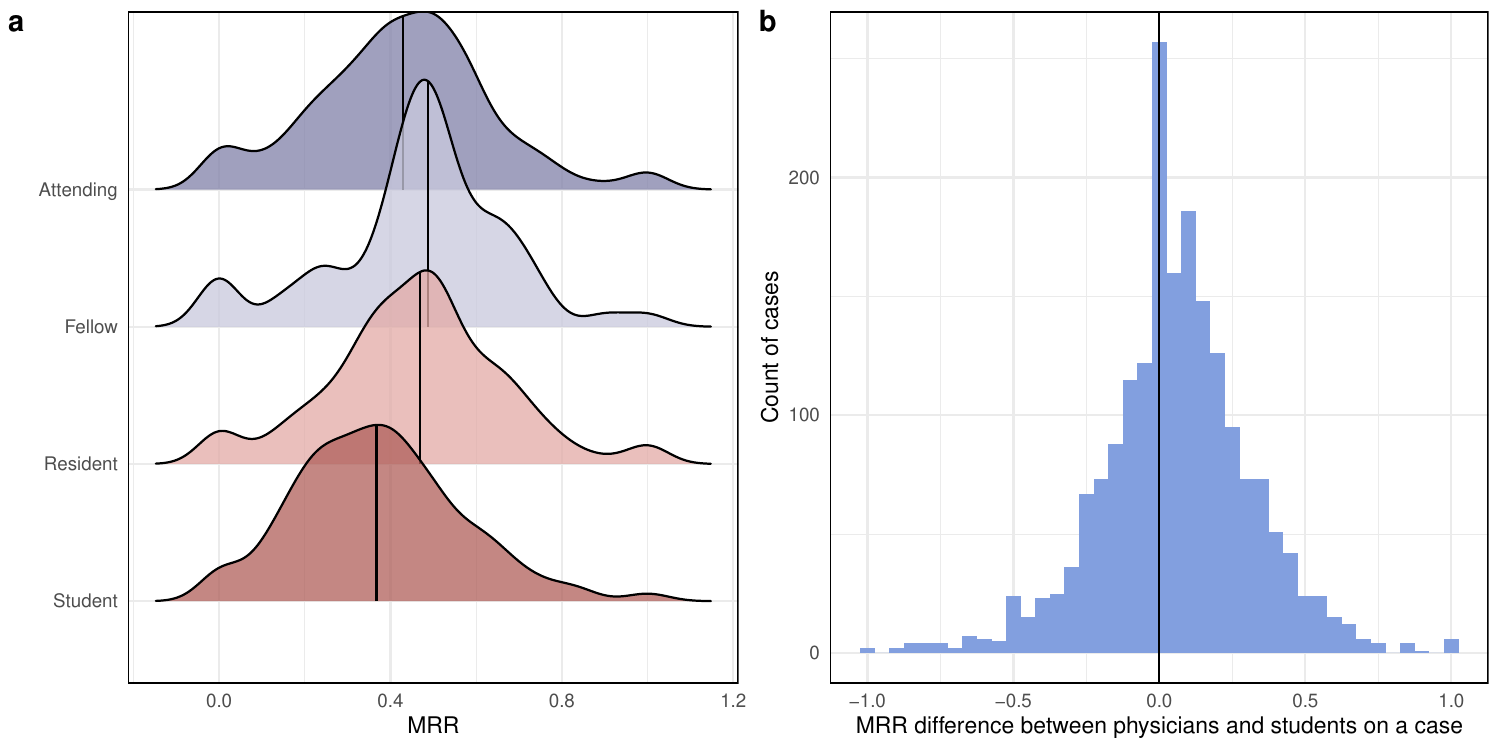}
\caption[Comparison of individual human performance across tenure levels.]{\textbf{Comparison of individual human performance across tenure levels.} \textbf{a}, Density estimates of individual MRR values ($x$ axis) for humans who solved at least five cases by tenure level. Vertical lines represent median MRR values within the tenure level. As the performance distributions of the three most experienced tenure levels (i.e., attending physicians, fellows, and resident physicians, representing senior doctors, doctors undergoing specialized training, and doctors in training, respectively) were similar, we combined these three groups. \textbf{b}, Comparing the pooled performance of physicians with that of medical students. For each case, the mean MRR across all physicians and all medical students was calculated. The distribution shows the difference in MRR between the two groups at the case level, with positive (negative) values indicating higher MRR for physicians (students). Physicians and students showed no difference in MRR in many cases, but the MRR for physicians was generally higher MRR than that for medical students.
}\label{fig:solvers_humans_expertise}
\end{figure}
\pagebreak

\begin{figure}[th]
\centering
\includegraphics[width=.9\textwidth]{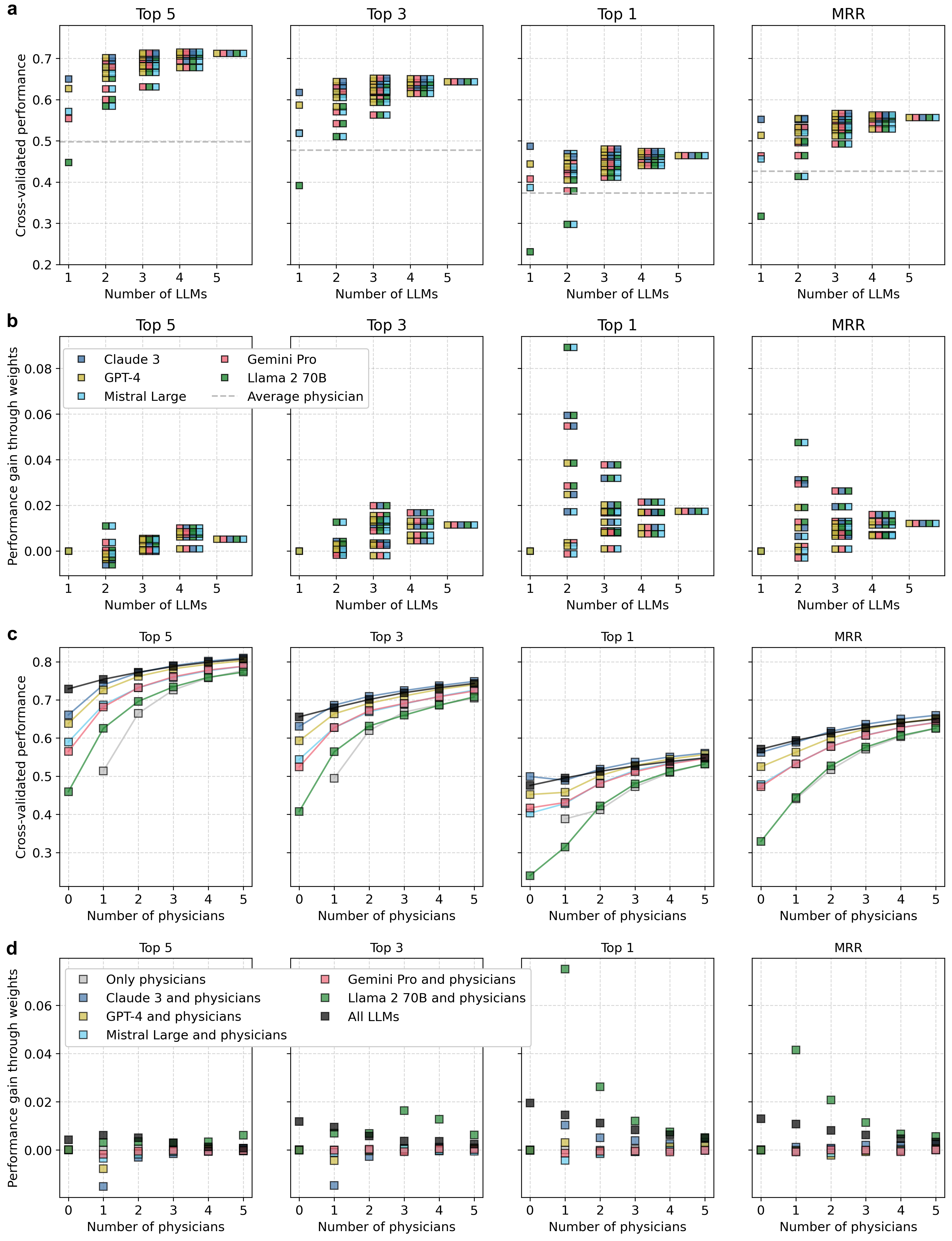}
\caption[Effect of weighting on cross-validated performance of LLM and hybrid human--LLM ensembles]{\textbf{Effect of weighting on cross-validated performance of LLM and hybrid human--LLM ensembles.} 
\textbf{a}, Cross-validated performance ($y$ axis) of five individual LLMs and ensembles of all possible combinations of LLMs with equal weights applied in the aggregation step. 
\textbf{b}, Signed difference in performance between weighted and unweighted aggregation ($y$ axis); higher values indicate greater gains from applying weights.
\textbf{c}, Cross-validated performance of hybrid physician-LLM ensembles with equal weights applied in the aggregation step ($y$ axis). 
\textbf{d}, Signed difference in performance between weighted and unweighted aggregation for hybrid human--LLM ensembles.
For most LLM- and hybrid ensembles, using the Weighted Majority Voting Ensemble approach \cite{Dogan2019} increased diagnostic accuracy, particularly for top-$1$ and MRR.
}\label{fig:llm-accuracy-weighted-vs-unweighted}
\end{figure}
\pagebreak

\FloatBarrier
\section{Supplementary information}

\begin{longtable}[c]{@{}lcccc@{}} 
    \caption[LLM ensemble performance]{LLM ensemble performance based on 2,133 case vignettes. Results reported are the averages of five-fold cross validation, where one fold was used to determine the best prompt and weights for each LLM according to WMVE (see Methods) and the other four folds were used to calculate performance.}    \label{tbl:LLMs}  \\
\toprule
Combo & Top 1 & Top 3 & Top 5 & MRR \\* \midrule
\endfirsthead
\multicolumn{5}{c}%
{{\bfseries Table \thetable\ continued from previous page}} \\
\toprule
Combo & Top 1 & Top 3 & Top 5 & MRR \\* \midrule
\endhead
\bottomrule
\endfoot
\bottomrule
\endlastfoot
Claude 3 & 0.487 & 0.618 & 0.650 & 0.553 \\
Claude 3|Llama 2 70B & 0.487 & 0.617 & 0.669 & 0.556 \\
Claude 3|Llama 2 70B|Mistral Large & 0.477 & 0.630 & 0.690 & 0.554 \\
Claude 3|Mistral Large & 0.487 & 0.628 & 0.684 & 0.558 \\
Gemini Pro & 0.408 & 0.519 & 0.554 & 0.464 \\
Gemini Pro|Claude 3 & 0.486 & 0.633 & 0.693 & 0.562 \\
Gemini Pro|Claude 3|Llama 2 70B & 0.473 & 0.639 & 0.696 & 0.558 \\
Gemini Pro|Claude 3|Llama 2 70B|Mistral Large & 0.469 & 0.645 & 0.705 & 0.557 \\
Gemini Pro|Claude 3|Mistral Large & 0.474 & 0.643 & 0.703 & 0.561 \\
Gemini Pro|Llama 2 70B & 0.408 & 0.541 & 0.601 & 0.477 \\
Gemini Pro|Llama 2 70B|Mistral Large & 0.421 & 0.572 & 0.637 & 0.500 \\
Gemini Pro|Mistral Large & 0.415 & 0.569 & 0.630 & 0.493 \\
Llama 2 70B & 0.231 & 0.392 & 0.448 & 0.317 \\
Llama 2 70B|Mistral Large & 0.387 & 0.523 & 0.596 & 0.461 \\
Mistral Large & 0.387 & 0.519 & 0.571 & 0.456 \\
GPT-4 & 0.444 & 0.587 & 0.627 & 0.514 \\
GPT-4|Claude 3 & 0.487 & 0.645 & 0.698 & 0.565 \\
GPT-4|Claude 3|Llama 2 70B & 0.486 & 0.643 & 0.706 & 0.565 \\
GPT-4|Claude 3|Llama 2 70B|Mistral Large & 0.481 & 0.647 & 0.712 & 0.565 \\
GPT-4|Claude 3|Mistral Large & 0.486 & 0.649 & 0.710 & 0.567 \\
GPT-4|Gemini Pro & 0.449 & 0.621 & 0.676 & 0.534 \\
GPT-4|Gemini Pro|Claude 3 & 0.488 & 0.655 & 0.713 & 0.573 \\
GPT-4|Gemini Pro|Claude 3|Llama 2 70B & 0.486 & 0.655 & 0.718 & 0.572 \\
GPT-4|Gemini Pro|Claude 3|Llama 2 70B|Mistral Large & 0.482 & 0.655 & 0.718 & 0.569 \\
GPT-4|Gemini Pro|Claude 3|Mistral Large & 0.484 & 0.655 & 0.717 & 0.570 \\
GPT-4|Gemini Pro|Llama 2 70B & 0.445 & 0.618 & 0.682 & 0.534 \\
GPT-4|Gemini Pro|Llama 2 70B|Mistral Large & 0.448 & 0.622 & 0.687 & 0.536 \\
GPT-4|Gemini Pro|Mistral Large & 0.447 & 0.619 & 0.683 & 0.535 \\
GPT-4|Llama 2 70B & 0.444 & 0.586 & 0.649 & 0.519 \\
GPT-4|Llama 2 70B|Mistral Large & 0.440 & 0.607 & 0.671 & 0.523 \\
GPT-4|Mistral Large & 0.439 & 0.606 & 0.663 & 0.520 \\
\end{longtable}


\begin{longtable}{lcccc}
\caption[LLM performance by prompt (detailed results)]{LLM performance per prompt and accuracy metric on the whole dataset of 2,133 cases. Note that in the cross-validation results presented in the main text, prompts were selected on a subset (one fold) of these cases.} \label{tbl:LLMs_prompting} \\

    \toprule
    \textbf{Prompt} & \textbf{MRR} & \textbf{Top 1} & \textbf{Top 3} & \textbf{Top 5} \\
    \midrule
    \endfirsthead

    \multicolumn{5}{c}{{\bfseries \tablename\ \thetable{} -- continued from previous page}} \\
    \toprule
    \textbf{Prompt} & \textbf{MRR} & \textbf{top1} & \textbf{top3} & \textbf{top5} \\
    \midrule
    \endhead

    \midrule \multicolumn{5}{r}{{Continued on next page}} \\
    \endfoot

    \bottomrule
    \endlastfoot

    \multicolumn{5}{c}{\textbf{Claude 3}} \\
    base\_common & 0.499 & 0.441 & 0.557 & 0.585 \\
    base\_common\_fewshot & 0.558 & 0.494 & 0.622 & 0.650 \\
    base\_common\_selfconsistent & 0.509 & 0.449 & 0.568 & 0.594 \\
    base\_common\_selfconsistent\_fewshot & 0.554 & 0.490 & 0.619 & 0.645 \\
    base\_impersonation\_common & 0.492 & 0.433 & 0.550 & 0.578 \\
    base\_impersonation\_common\_fewshot & 0.553 & 0.488 & 0.617 & 0.646 \\
    base\_impersonation\_common\_selfconsistent & 0.501 & 0.440 & 0.562 & 0.586 \\
    base\_impersonation\_common\_selfconsistent\_fewshot & 0.557 & 0.492 & 0.622 & 0.651 \\
    base\_impersonation\_sct & 0.454 & 0.379 & 0.526 & 0.568 \\
    base\_impersonation\_sct\_fewshot & 0.543 & 0.471 & 0.607 & 0.654 \\
    base\_impersonation\_sct\_selfconsistent & 0.469 & 0.396 & 0.542 & 0.579 \\
    base\_impersonation\_sct\_selfconsistent\_fewshot & 0.539 & 0.465 & 0.608 & 0.649 \\
    base\_sct & 0.453 & 0.379 & 0.525 & 0.565 \\
    base\_sct\_fewshot & 0.549 & 0.478 & 0.620 & 0.658 \\
    base\_sct\_selfconsistent & 0.460 & 0.388 & 0.529 & 0.571 \\
    base\_sct\_selfconsistent\_fewshot & 0.542 & 0.468 & 0.613 & 0.651 \\
    \midrule
    \multicolumn{5}{c}{\textbf{Gemini Pro}} \\
    base\_common & 0.386 & 0.339 & 0.430 & 0.458 \\
    base\_common\_fewshot & 0.462 & 0.411 & 0.508 & 0.539 \\
    base\_common\_selfconsistent & 0.391 & 0.346 & 0.432 & 0.460 \\
    base\_common\_selfconsistent\_fewshot & 0.466 & 0.413 & 0.513 & 0.548 \\
    base\_impersonation\_common & 0.391 & 0.343 & 0.436 & 0.463 \\
    base\_impersonation\_common\_fewshot & 0.450 & 0.395 & 0.499 & 0.533 \\
    base\_impersonation\_common\_selfconsistent & 0.398 & 0.350 & 0.443 & 0.470 \\
    base\_impersonation\_common\_selfconsistent\_fewshot & 0.457 & 0.401 & 0.505 & 0.545 \\
    base\_impersonation\_sct & 0.402 & 0.328 & 0.472 & 0.515 \\
    base\_impersonation\_sct\_fewshot & 0.465 & 0.407 & 0.519 & 0.557 \\
    base\_impersonation\_sct\_selfconsistent & 0.413 & 0.341 & 0.482 & 0.522 \\
    base\_impersonation\_sct\_selfconsistent\_fewshot & 0.465 & 0.403 & 0.523 & 0.559 \\
    base\_sct & 0.406 & 0.335 & 0.472 & 0.512 \\
    base\_sct\_fewshot & 0.470 & 0.412 & 0.523 & 0.559 \\
    base\_sct\_selfconsistent & 0.423 & 0.354 & 0.487 & 0.529 \\
    base\_sct\_selfconsistent\_fewshot & 0.468 & 0.408 & 0.523 & 0.558 \\
    \midrule
    \multicolumn{5}{c}{\textbf{Llama 2 70B}} \\
    base\_common & 0.311 & 0.225 & 0.387 & 0.446 \\
    base\_common\_fewshot & 0.271 & 0.228 & 0.305 & 0.339 \\
    base\_common\_selfconsistent & 0.315 & 0.231 & 0.389 & 0.450 \\
    base\_common\_selfconsistent\_fewshot & 0.285 & 0.224 & 0.335 & 0.377 \\
    base\_impersonation\_common & 0.296 & 0.211 & 0.373 & 0.431 \\
    base\_impersonation\_common\_fewshot & 0.278 & 0.217 & 0.331 & 0.372 \\
    base\_impersonation\_common\_selfconsistent & 0.303 & 0.215 & 0.383 & 0.440 \\
    base\_impersonation\_common\_selfconsistent\_fewshot & 0.284 & 0.208 & 0.348 & 0.401 \\
    base\_impersonation\_sct & 0.305 & 0.221 & 0.381 & 0.438 \\
    base\_impersonation\_sct\_fewshot & 0.231 & 0.146 & 0.310 & 0.364 \\
    base\_impersonation\_sct\_selfconsistent & 0.312 & 0.228 & 0.388 & 0.446 \\
    base\_impersonation\_sct\_selfconsistent\_fewshot & 0.246 & 0.154 & 0.332 & 0.386 \\
    base\_sct & 0.312 & 0.228 & 0.386 & 0.446 \\
    base\_sct\_fewshot & 0.228 & 0.164 & 0.287 & 0.331 \\
    base\_sct\_selfconsistent & 0.323 & 0.238 & 0.401 & 0.455 \\
    base\_sct\_selfconsistent\_fewshot & 0.263 & 0.184 & 0.335 & 0.386 \\
    \midrule
    \multicolumn{5}{c}{\textbf{Mistral Large}} \\
    base\_common & 0.428 & 0.347 & 0.505 & 0.546 \\
    base\_common\_fewshot & 0.468 & 0.395 & 0.532 & 0.581 \\
    base\_common\_selfconsistent & 0.434 & 0.353 & 0.511 & 0.550 \\
    base\_common\_selfconsistent\_fewshot & 0.467 & 0.395 & 0.532 & 0.580 \\
    base\_impersonation\_common & 0.428 & 0.356 & 0.493 & 0.536 \\
    base\_impersonation\_common\_fewshot & 0.460 & 0.390 & 0.525 & 0.570 \\
    base\_impersonation\_common\_selfconsistent & 0.428 & 0.358 & 0.493 & 0.532 \\
    base\_impersonation\_common\_selfconsistent\_fewshot & 0.463 & 0.391 & 0.528 & 0.577 \\
    base\_impersonation\_sct & 0.321 & 0.237 & 0.400 & 0.451 \\
    base\_impersonation\_sct\_fewshot & 0.445 & 0.373 & 0.505 & 0.561 \\
    base\_impersonation\_sct\_selfconsistent & 0.326 & 0.244 & 0.409 & 0.450 \\
    base\_impersonation\_sct\_selfconsistent\_fewshot & 0.446 & 0.372 & 0.508 & 0.566 \\
    base\_sct & 0.347 & 0.256 & 0.435 & 0.484 \\
    base\_sct\_fewshot & 0.447 & 0.372 & 0.512 & 0.568 \\
    base\_sct\_selfconsistent & 0.355 & 0.261 & 0.442 & 0.498 \\
    base\_sct\_selfconsistent\_fewshot & 0.449 & 0.374 & 0.515 & 0.570 \\
    \midrule
    \multicolumn{5}{c}{\textbf{GPT-4}} \\
    base\_common & 0.478 & 0.412 & 0.541 & 0.577 \\
    base\_common\_fewshot & 0.519 & 0.451 & 0.584 & 0.616 \\
    base\_common\_selfconsistent & 0.487 & 0.423 & 0.548 & 0.580 \\
    base\_common\_selfconsistent\_fewshot & 0.516 & 0.446 & 0.584 & 0.618 \\
    base\_impersonation\_common & 0.462 & 0.396 & 0.527 & 0.557 \\
    base\_impersonation\_common\_fewshot & 0.503 & 0.434 & 0.570 & 0.600 \\
    base\_impersonation\_common\_selfconsistent & 0.471 & 0.409 & 0.531 & 0.562 \\
    base\_impersonation\_common\_selfconsistent\_fewshot & 0.509 & 0.443 & 0.575 & 0.606 \\
    base\_impersonation\_sct & 0.410 & 0.329 & 0.486 & 0.531 \\
    base\_impersonation\_sct\_fewshot & 0.504 & 0.425 & 0.575 & 0.623 \\
    base\_impersonation\_sct\_selfconsistent & 0.433 & 0.356 & 0.502 & 0.544 \\
    base\_impersonation\_sct\_selfconsistent\_fewshot & 0.512 & 0.433 & 0.584 & 0.628 \\
    base\_sct & 0.418 & 0.338 & 0.491 & 0.537 \\
    base\_sct\_fewshot & 0.516 & 0.440 & 0.587 & 0.630 \\
    base\_sct\_selfconsistent & 0.443 & 0.367 & 0.514 & 0.555 \\
    base\_sct\_selfconsistent\_fewshot & 0.517 & 0.438 & 0.592 & 0.631 \\
\end{longtable}

\subsection{Example case vignette}

\textbf{40 year old female presenting with chest pain} \\
\noindent \textit{Symptom:} Worsening chest pain
\begin{itemize}
    \item Onset: 2 weeks ago
    \item Associated with: Cough, dyspnea, fever
    \item Complicated by: Fatigue
\end{itemize}

\noindent \textit{Social history}
\begin{itemize}
\item  Recent construction in Ohio
\end{itemize}

\noindent \textit{Physical exam}
\begin{itemize}
\item  Lungs: Wheezing
\end{itemize}

\noindent \textit{Diagnostic: X-ray}
\begin{itemize}
\item  Interpretation: Normal
\end{itemize}

\subsection{Verbatim text of LLM prompts}
\label{sec:verbatim-llm-prompts}
Below we list the text of the building blocks used to construct prompts for eliciting differential diagnoses. For all LLMs, we tested all feasible combinations of these building blocks. Diagnostic accuracies for each LLM and each prompt combination on the whole dataset of 2133 cases are presented in Extended Data Table \ref{tbl:LLMs_prompting}.

\noindent \textbf{Base task}: \textit{ Provide only the most probable differential diagnosis, no explanation, no recapitulation of the case information or task. Give a maximum of 5 answers, sorted by probability of being the correct diagnosis, most probable first, remove list numbering, and respond with each answer on a new line. Be as concise as possible, no need to be polite. }
                 
\noindent\textbf{Impersonation}: \textit{You are a medical expert diagnosing a patient.}

\noindent\textbf{Response format: common shorthand}: \textit{In your answer use common shorthand non-abbreviated diagnoses.}

\noindent\textbf{Response format: SCT}: \textit{In your answer provide only the appropriate SNOMED CT fully specified name, no id.}

\noindent\textbf{Self-consistency}: \textit{Check that each differential diagnosis in your answer is consistent with each finding in the case description. }

\noindent\textbf{Few-shot prompt}: \textit{Here are some examples of cases and their correct answers:} \\
\textit{Case description: \{case vignette\}} \\
\textit{Answer: \{example solution\} }(5x) \\

\vspace{10pt}
\noindent The prompt used to have \llmClaude \, assign a medical specialty to a case vignette was as follows: \\
\textit{In the list of medical specialties within the following \textless SP\textgreater \textbackslash\textless SP\textgreater tags \textless SP\textgreater\{specialties\_one\_line\}\textless\textbackslash SP\textgreater, each specialty is separated by a comma. You are given the following case in the \textless case\textgreater \textbackslash\textless case\textgreater tags: \textless case\textgreater\{case\_text\}\textless\textbackslash case\textgreater. Follow the steps provided: 1. Determine the top 3 choices for which specialties best fit the case. 2. Give your answer in an ordered numbered list starting with the most confident answer first. Only answer with the list. Do not provide any additional explanation.}

\end{appendices}
\pagebreak
\bibliography{__refs}


\begin{thebibliography}{79}
\ifx \bisbn   \undefined \def \bisbn  #1{ISBN #1}\fi
\ifx \binits  \undefined \def \binits#1{#1}\fi
\ifx \bauthor  \undefined \def \bauthor#1{#1}\fi
\ifx \batitle  \undefined \def \batitle#1{#1}\fi
\ifx \bjtitle  \undefined \def \bjtitle#1{#1}\fi
\ifx \bvolume  \undefined \def \bvolume#1{\textbf{#1}}\fi
\ifx \byear  \undefined \def \byear#1{#1}\fi
\ifx \bissue  \undefined \def \bissue#1{#1}\fi
\ifx \bfpage  \undefined \def \bfpage#1{#1}\fi
\ifx \blpage  \undefined \def \blpage #1{#1}\fi
\ifx \burl  \undefined \def \burl#1{\textsf{#1}}\fi
\ifx \doiurl  \undefined \def \doiurl#1{\url{https://doi.org/#1}}\fi
\ifx \betal  \undefined \def \betal{\textit{et al.}}\fi
\ifx \binstitute  \undefined \def \binstitute#1{#1}\fi
\ifx \binstitutionaled  \undefined \def \binstitutionaled#1{#1}\fi
\ifx \bctitle  \undefined \def \bctitle#1{#1}\fi
\ifx \beditor  \undefined \def \beditor#1{#1}\fi
\ifx \bpublisher  \undefined \def \bpublisher#1{#1}\fi
\ifx \bbtitle  \undefined \def \bbtitle#1{#1}\fi
\ifx \bedition  \undefined \def \bedition#1{#1}\fi
\ifx \bseriesno  \undefined \def \bseriesno#1{#1}\fi
\ifx \blocation  \undefined \def \blocation#1{#1}\fi
\ifx \bsertitle  \undefined \def \bsertitle#1{#1}\fi
\ifx \bsnm \undefined \def \bsnm#1{#1}\fi
\ifx \bsuffix \undefined \def \bsuffix#1{#1}\fi
\ifx \bparticle \undefined \def \bparticle#1{#1}\fi
\ifx \barticle \undefined \def \barticle#1{#1}\fi
\bibcommenthead
\ifx \bconfdate \undefined \def \bconfdate #1{#1}\fi
\ifx \botherref \undefined \def \botherref #1{#1}\fi
\ifx \url \undefined \def \url#1{\textsf{#1}}\fi
\ifx \bchapter \undefined \def \bchapter#1{#1}\fi
\ifx \bbook \undefined \def \bbook#1{#1}\fi
\ifx \bcomment \undefined \def \bcomment#1{#1}\fi
\ifx \oauthor \undefined \def \oauthor#1{#1}\fi
\ifx \citeauthoryear \undefined \def \citeauthoryear#1{#1}\fi
\ifx \endbibitem  \undefined \def \endbibitem {}\fi
\ifx \bconflocation  \undefined \def \bconflocation#1{#1}\fi
\ifx \arxivurl  \undefined \def \arxivurl#1{\textsf{#1}}\fi
\csname PreBibitemsHook\endcsname

\bibitem[\protect\citeauthoryear{Hong et~al.}{2024}]{hong2024hallucinations}
\begin{botherref}
\oauthor{\bsnm{Hong}, \binits{G.}},
\oauthor{\bsnm{Gema}, \binits{A.P.}},
\oauthor{\bsnm{Saxena}, \binits{R.}},
\oauthor{\bsnm{Du}, \binits{X.}},
\oauthor{\bsnm{Nie}, \binits{P.}},
\oauthor{\bsnm{Zhao}, \binits{Y.}},
\oauthor{\bsnm{Perez-Beltrachini}, \binits{L.}},
\oauthor{\bsnm{Ryabinin}, \binits{M.}},
\oauthor{\bsnm{He}, \binits{X.}},
\oauthor{\bsnm{Fourrier}, \binits{C.}},
\oauthor{\bsnm{Minervini}, \binits{P.}}:
The Hallucinations Leaderboard: An open effort to measure hallucinations in large language models.
arXiv
(2024).
\url{https://arxiv.org/html/2404.05904v1}
\end{botherref}
\endbibitem

\bibitem[\protect\citeauthoryear{Ji et~al.}{2023}]{Ji2023}
\begin{barticle}
\bauthor{\bsnm{Ji}, \binits{Z.}},
\bauthor{\bsnm{Lee}, \binits{N.}},
\bauthor{\bsnm{Frieske}, \binits{R.}},
\bauthor{\bsnm{Yu}, \binits{T.}},
\bauthor{\bsnm{Su}, \binits{D.}},
\bauthor{\bsnm{Xu}, \binits{Y.}},
\bauthor{\bsnm{Ishii}, \binits{E.}},
\bauthor{\bsnm{Bang}, \binits{Y.J.}},
\bauthor{\bsnm{Madotto}, \binits{A.}},
\bauthor{\bsnm{Fung}, \binits{P.}}:
\batitle{Survey of hallucination in natural language generation}.
\bjtitle{ACM Computing Surveys}
\bvolume{55}(\bissue{12}),
\bfpage{1}--\blpage{38}
(\byear{2023})
\doiurl{10.1145/3571730}
\end{barticle}
\endbibitem

\bibitem[\protect\citeauthoryear{Pal et~al.}{2023}]{Pal2023}
\begin{botherref}
\oauthor{\bsnm{Pal}, \binits{A.}},
\oauthor{\bsnm{Umapathi}, \binits{L.K.}},
\oauthor{\bsnm{Sankarasubbu}, \binits{M.}}:
Med-HALT: Medical Domain Hallucination Test for large language models.
arXiv
(2023).
\url{https://arxiv.org/abs/2307.15343}
\end{botherref}
\endbibitem

\bibitem[\protect\citeauthoryear{Tonmoy et~al.}{2024}]{tonmoy2024Comprehensive}
\begin{botherref}
\oauthor{\bsnm{Tonmoy}, \binits{S.M.T.I.}},
\oauthor{\bsnm{Zaman}, \binits{S.M.M.}},
\oauthor{\bsnm{Jain}, \binits{V.}},
\oauthor{\bsnm{Rani}, \binits{A.}},
\oauthor{\bsnm{Rawte}, \binits{V.}},
\oauthor{\bsnm{Chadha}, \binits{A.}},
\oauthor{\bsnm{Das}, \binits{A.}}:
A comprehensive survey of hallucination mitigation techniques in large language models.
arXiv
(2024).
\url{http://arxiv.org/abs/2401.01313}
\end{botherref}
\endbibitem

\bibitem[\protect\citeauthoryear{Williams and Huckle}{2024}]{williams2024easy}
\begin{botherref}
\oauthor{\bsnm{Williams}, \binits{S.}},
\oauthor{\bsnm{Huckle}, \binits{J.}}:
Easy problems that LLMs get wrong.
arXiv
(2024).
\url{https://arxiv.org/html/2405.19616}
\end{botherref}
\endbibitem

\bibitem[\protect\citeauthoryear{Omiye et~al.}{2023}]{Omiye2023}
\begin{barticle}
\bauthor{\bsnm{Omiye}, \binits{J.A.}},
\bauthor{\bsnm{Lester}, \binits{J.C.}},
\bauthor{\bsnm{Spichak}, \binits{S.}},
\bauthor{\bsnm{Rotemberg}, \binits{V.}},
\bauthor{\bsnm{Daneshjou}, \binits{R.}}:
\batitle{Large language models propagate race-based medicine}.
\bjtitle{npj Digital Medicine}
\bvolume{6}(\bissue{1}),
\bfpage{195}
(\byear{2023})
\doiurl{10.1038/s41746-023-00939-z}
\end{barticle}
\endbibitem

\bibitem[\protect\citeauthoryear{Navigli et~al.}{2023}]{Navigli2023}
\begin{barticle}
\bauthor{\bsnm{Navigli}, \binits{R.}},
\bauthor{\bsnm{Conia}, \binits{S.}},
\bauthor{\bsnm{Ross}, \binits{B.}}:
\batitle{Biases in large language models: Origins, inventory, and discussion}.
\bjtitle{Journal of Data and Information Quality}
\bvolume{15}(\bissue{2}),
\bfpage{1}--\blpage{21}
(\byear{2023})
\doiurl{10.1145/3597307}
\end{barticle}
\endbibitem

\bibitem[\protect\citeauthoryear{Trianni et~al.}{2023}]{Trianni:HHAI2023}
\begin{bchapter}
\bauthor{\bsnm{Trianni}, \binits{V.}},
\bauthor{\bsnm{Nuzzolese}, \binits{A.G.}},
\bauthor{\bsnm{Porciello}, \binits{J.}},
\bauthor{\bsnm{Kurvers}, \binits{R.H.J.M.}},
\bauthor{\bsnm{Herzog}, \binits{S.M.}},
\bauthor{\bsnm{Barabucci}, \binits{G.}},
\bauthor{\bsnm{Berditchevskaia}, \binits{A.}},
\bauthor{\bsnm{Fung}, \binits{F.}}:
\bctitle{Hybrid collective intelligence for decision support in complex open-ended domains}.
In: \beditor{\bsnm{Lukowicz}, \binits{P.}},
\beditor{\bsnm{Mayer}, \binits{S.}},
\beditor{\bsnm{Koch}, \binits{J.}},
\beditor{\bsnm{Shawe-Taylor}, \binits{J.}},
\beditor{\bsnm{Tiddi}, \binits{I.}} (eds.)
\bbtitle{{HHAI 2023: Augmenting Human Intellect}}.
\bsertitle{Frontiers in Artificial Intelligence and Applications},
vol. \bseriesno{368},
pp. \bfpage{124}--\blpage{137}.
\bpublisher{IOS Press},
\blocation{Amsterdam, Netherlands}
(\byear{2023}).
\doiurl{10.3233/FAIA230079}
\end{bchapter}
\endbibitem

\bibitem[\protect\citeauthoryear{Makary and Daniel}{2016}]{makary2016medical}
\begin{barticle}
\bauthor{\bsnm{Makary}, \binits{M.A.}},
\bauthor{\bsnm{Daniel}, \binits{M.}}:
\batitle{Medical error: The third leading cause of death in the {US}}.
\bjtitle{BMJ}
\bvolume{353},
\bfpage{2139}
(\byear{2016})
\doiurl{10.1136/bmj.i2139}
\end{barticle}
\endbibitem

\bibitem[\protect\citeauthoryear{Leape et~al.}{1991}]{leape1991nature}
\begin{barticle}
\bauthor{\bsnm{Leape}, \binits{L.L.}},
\bauthor{\bsnm{Brennan}, \binits{T.A.}},
\bauthor{\bsnm{Laird}, \binits{N.}},
\bauthor{\bsnm{Lawthers}, \binits{A.G.}},
\bauthor{\bsnm{Localio}, \binits{A.R.}},
\bauthor{\bsnm{Barnes}, \binits{B.A.}},
\bauthor{\bsnm{Hebert}, \binits{L.}},
\bauthor{\bsnm{Newhouse}, \binits{J.P.}},
\bauthor{\bsnm{Weiler}, \binits{P.C.}},
\bauthor{\bsnm{Hiatt}, \binits{H.}}:
\batitle{The nature of adverse events in hospitalized patients: {Results of the Harvard Medical Practice Study II}}.
\bjtitle{New England Journal of Medicine}
\bvolume{324}(\bissue{6}),
\bfpage{377}--\blpage{384}
(\byear{1991})
\doiurl{10.1056/NEJM199102073240605}
\end{barticle}
\endbibitem

\bibitem[\protect\citeauthoryear{Graber et~al.}{2005}]{graber2005diagnostic}
\begin{barticle}
\bauthor{\bsnm{Graber}, \binits{M.L.}},
\bauthor{\bsnm{Franklin}, \binits{N.}},
\bauthor{\bsnm{Gordon}, \binits{R.}}:
\batitle{Diagnostic error in internal medicine}.
\bjtitle{Archives of Internal Medicine}
\bvolume{165}(\bissue{13}),
\bfpage{1493}--\blpage{1499}
(\byear{2005})
\doiurl{10.1001/archinte.165.13.1493}
\end{barticle}
\endbibitem

\bibitem[\protect\citeauthoryear{Newman-Toker et~al.}{2024}]{newman2024}
\begin{barticle}
\bauthor{\bsnm{Newman-Toker}, \binits{D.E.}},
\bauthor{\bsnm{Nassery}, \binits{N.}},
\bauthor{\bsnm{Schaffer}, \binits{A.C.}},
\bauthor{\bsnm{Yu-Moe}, \binits{C.W.}},
\bauthor{\bsnm{Clemens}, \binits{G.D.}},
\bauthor{\bsnm{Wang}, \binits{Z.}},
\bauthor{\bsnm{Zhu}, \binits{Y.}},
\bauthor{\bsnm{Tehrani}, \binits{A.S.S.}},
\bauthor{\bsnm{Fanai}, \binits{M.}},
\bauthor{\bsnm{Hassoon}, \binits{A.}},
\bauthor{\bsnm{Siegal}, \binits{D.}}:
\batitle{Burden of serious harms from diagnostic error in the {USA}}.
\bjtitle{BMJ Quality \& Safety}
\bvolume{33}(\bissue{2}),
\bfpage{109}--\blpage{120}
(\byear{2024})
\doiurl{10.1136/bmjqs-2021-014130}
\end{barticle}
\endbibitem

\bibitem[\protect\citeauthoryear{Basu et~al.}{2020}]{basu2020artificial}
\begin{barticle}
\bauthor{\bsnm{Basu}, \binits{K.}},
\bauthor{\bsnm{Sinha}, \binits{R.}},
\bauthor{\bsnm{Ong}, \binits{A.}},
\bauthor{\bsnm{Basu}, \binits{T.}}:
\batitle{Artificial intelligence: How is it changing medical sciences and its future?}
\bjtitle{Indian Journal of Dermatology}
\bvolume{65}(\bissue{5}),
\bfpage{365}--\blpage{370}
(\byear{2020})
\doiurl{10.4103/ijd.IJD_421_20}
\end{barticle}
\endbibitem

\bibitem[\protect\citeauthoryear{Mirbabaie et~al.}{2021}]{mirbabaie2021artificial}
\begin{barticle}
\bauthor{\bsnm{Mirbabaie}, \binits{M.}},
\bauthor{\bsnm{Stieglitz}, \binits{S.}},
\bauthor{\bsnm{Frick}, \binits{N.R.}}:
\batitle{Artificial intelligence in disease diagnostics: A critical review and classification on the current state of research guiding future direction}.
\bjtitle{Health and Technology}
\bvolume{11}(\bissue{4}),
\bfpage{693}--\blpage{731}
(\byear{2021})
\doiurl{10.1007/s12553-021-00555-5}
\end{barticle}
\endbibitem

\bibitem[\protect\citeauthoryear{Rajpurkar et~al.}{2022}]{rajpurkar2022ai}
\begin{barticle}
\bauthor{\bsnm{Rajpurkar}, \binits{P.}},
\bauthor{\bsnm{Chen}, \binits{E.}},
\bauthor{\bsnm{Banerjee}, \binits{O.}},
\bauthor{\bsnm{Topol}, \binits{E.J.}}:
\batitle{{AI} in health and medicine}.
\bjtitle{Nature Medicine}
\bvolume{28}(\bissue{1}),
\bfpage{31}--\blpage{38}
(\byear{2022})
\doiurl{10.1038/s41591-021-01614-0}
\end{barticle}
\endbibitem

\bibitem[\protect\citeauthoryear{Aggarwal et~al.}{2021}]{aggarwal2021diagnostic}
\begin{barticle}
\bauthor{\bsnm{Aggarwal}, \binits{R.}},
\bauthor{\bsnm{Sounderajah}, \binits{V.}},
\bauthor{\bsnm{Martin}, \binits{G.}},
\bauthor{\bsnm{Ting}, \binits{D.S.W.}},
\bauthor{\bsnm{Karthikesalingam}, \binits{A.}},
\bauthor{\bsnm{King}, \binits{D.}},
\bauthor{\bsnm{Ashrafian}, \binits{H.}},
\bauthor{\bsnm{Darzi}, \binits{A.}}:
\batitle{Diagnostic accuracy of deep learning in medical imaging: A systematic review and meta-analysis}.
\bjtitle{npj Digital Medicine}
\bvolume{4}(\bissue{1}),
\bfpage{65}
(\byear{2021})
\doiurl{10.1038/s41746-021-00438-z}
\end{barticle}
\endbibitem

\bibitem[\protect\citeauthoryear{Dembrower et~al.}{2023}]{dembrowerArtificialIntelligenceBreast2023}
\begin{barticle}
\bauthor{\bsnm{Dembrower}, \binits{K.}},
\bauthor{\bsnm{Crippa}, \binits{A.}},
\bauthor{\bsnm{Col{\'o}n}, \binits{E.}},
\bauthor{\bsnm{Eklund}, \binits{M.}},
\bauthor{\bsnm{Strand}, \binits{F.}}:
\batitle{Artificial intelligence for breast cancer detection in screening mammography in {{Sweden}}: A prospective, population-based, paired-reader, non-inferiority study}.
\bjtitle{The Lancet Digital Health}
\bvolume{5}(\bissue{10}),
\bfpage{703}--\blpage{711}
(\byear{2023})
\doiurl{10.1016/S2589-7500(23)00153-X}
\end{barticle}
\endbibitem

\bibitem[\protect\citeauthoryear{Lu et~al.}{2024}]{lu2024multimodal}
\begin{botherref}
\oauthor{\bsnm{Lu}, \binits{M.Y.}},
\oauthor{\bsnm{Chen}, \binits{B.}},
\oauthor{\bsnm{Williamson}, \binits{D.F.K.}},
\oauthor{\bsnm{Chen}, \binits{R.J.}},
\oauthor{\bsnm{Zhao}, \binits{M.}},
\oauthor{\bsnm{Chow}, \binits{A.K.}},
\oauthor{\bsnm{Ikemura}, \binits{K.}},
\oauthor{\bsnm{Kim}, \binits{A.}},
\oauthor{\bsnm{Pouli}, \binits{D.}},
\oauthor{\bsnm{Patel}, \binits{A.}},
\oauthor{\bsnm{Soliman}, \binits{A.}},
\oauthor{\bsnm{Chen}, \binits{C.}},
\oauthor{\bsnm{Ding}, \binits{T.}},
\oauthor{\bsnm{Wang}, \binits{J.J.}},
\oauthor{\bsnm{Gerber}, \binits{G.}},
\oauthor{\bsnm{Liang}, \binits{I.}},
\oauthor{\bsnm{Le}, \binits{L.P.}},
\oauthor{\bsnm{Parwani}, \binits{A.V.}},
\oauthor{\bsnm{Weishaupt}, \binits{L.L.}},
\oauthor{\bsnm{Mahmood}, \binits{F.}}:
A multimodal generative {{AI}} copilot for human pathology.
Nature,
1--3
(2024)
\doiurl{10.1038/s41586-024-07618-3} .
Accessed 2024-06-14
\end{botherref}
\endbibitem

\bibitem[\protect\citeauthoryear{Moor et~al.}{2023}]{moorFoundationModelsGeneralist2023}
\begin{barticle}
\bauthor{\bsnm{Moor}, \binits{M.}},
\bauthor{\bsnm{Banerjee}, \binits{O.}},
\bauthor{\bsnm{Abad}, \binits{Z.S.H.}},
\bauthor{\bsnm{Krumholz}, \binits{H.M.}},
\bauthor{\bsnm{Leskovec}, \binits{J.}},
\bauthor{\bsnm{Topol}, \binits{E.J.}},
\bauthor{\bsnm{Rajpurkar}, \binits{P.}}:
\batitle{Foundation models for generalist medical artificial intelligence}.
\bjtitle{Nature}
\bvolume{616}(\bissue{7956}),
\bfpage{259}--\blpage{265}
(\byear{2023})
\doiurl{10.1038/s41586-023-05881-4}
\end{barticle}
\endbibitem

\bibitem[\protect\citeauthoryear{Singhal et~al.}{2023}]{singhalLargeLanguageModels2023}
\begin{barticle}
\bauthor{\bsnm{Singhal}, \binits{K.}},
\bauthor{\bsnm{Azizi}, \binits{S.}},
\bauthor{\bsnm{Tu}, \binits{T.}},
\bauthor{\bsnm{Mahdavi}, \binits{S.S.}},
\bauthor{\bsnm{Wei}, \binits{J.}},
\bauthor{\bsnm{Chung}, \binits{H.W.}},
\bauthor{\bsnm{Scales}, \binits{N.}},
\bauthor{\bsnm{Tanwani}, \binits{A.}},
\bauthor{\bsnm{{Cole-Lewis}}, \binits{H.}},
\bauthor{\bsnm{Pfohl}, \binits{S.}},
\bauthor{\bsnm{Payne}, \binits{P.}},
\bauthor{\bsnm{Seneviratne}, \binits{M.}},
\bauthor{\bsnm{Gamble}, \binits{P.}},
\bauthor{\bsnm{Kelly}, \binits{C.}},
\bauthor{\bsnm{Babiker}, \binits{A.}},
\bauthor{\bsnm{Sch{\"a}rli}, \binits{N.}},
\bauthor{\bsnm{Chowdhery}, \binits{A.}},
\bauthor{\bsnm{Mansfield}, \binits{P.}},
\bauthor{\bsnm{{Demner-Fushman}}, \binits{D.}},
\bauthor{\bsnm{{Ag{\"u}era y Arcas}}, \binits{B.}},
\bauthor{\bsnm{Webster}, \binits{D.}},
\bauthor{\bsnm{Corrado}, \binits{G.S.}},
\bauthor{\bsnm{Matias}, \binits{Y.}},
\bauthor{\bsnm{Chou}, \binits{K.}},
\bauthor{\bsnm{Gottweis}, \binits{J.}},
\bauthor{\bsnm{Tomasev}, \binits{N.}},
\bauthor{\bsnm{Liu}, \binits{Y.}},
\bauthor{\bsnm{Rajkomar}, \binits{A.}},
\bauthor{\bsnm{Barral}, \binits{J.}},
\bauthor{\bsnm{Semturs}, \binits{C.}},
\bauthor{\bsnm{Karthikesalingam}, \binits{A.}},
\bauthor{\bsnm{Natarajan}, \binits{V.}}:
\batitle{Large language models encode clinical knowledge}.
\bjtitle{Nature}
\bvolume{620}(\bissue{7972}),
\bfpage{172}--\blpage{180}
(\byear{2023})
\doiurl{10.1038/s41586-023-06291-2}
\end{barticle}
\endbibitem

\bibitem[\protect\citeauthoryear{Jiang et~al.}{2023}]{Jiang2023}
\begin{barticle}
\bauthor{\bsnm{Jiang}, \binits{L.Y.}},
\bauthor{\bsnm{Liu}, \binits{X.C.}},
\bauthor{\bsnm{Nejatian}, \binits{N.P.}},
\bauthor{\bsnm{Nasir-Moin}, \binits{M.}},
\bauthor{\bsnm{Wang}, \binits{D.}},
\bauthor{\bsnm{Abidin}, \binits{A.}},
\bauthor{\bsnm{Eaton}, \binits{K.}},
\bauthor{\bsnm{Riina}, \binits{H.A.}},
\bauthor{\bsnm{Laufer}, \binits{I.}},
\bauthor{\bsnm{Punjabi}, \binits{P.}},
\bauthor{\bsnm{Miceli}, \binits{M.}},
\bauthor{\bsnm{Kim}, \binits{N.C.}},
\bauthor{\bsnm{Orillac}, \binits{C.}},
\bauthor{\bsnm{Schnurman}, \binits{Z.}},
\bauthor{\bsnm{Livia}, \binits{C.}},
\bauthor{\bsnm{Weiss}, \binits{H.}},
\bauthor{\bsnm{Kurland}, \binits{D.}},
\bauthor{\bsnm{Neifert}, \binits{S.}},
\bauthor{\bsnm{Dastagirzada}, \binits{Y.}},
\bauthor{\bsnm{Kondziolka}, \binits{D.}},
\bauthor{\bsnm{Cheung}, \binits{A.T.M.}},
\bauthor{\bsnm{Yang}, \binits{G.}},
\bauthor{\bsnm{Cao}, \binits{M.}},
\bauthor{\bsnm{Flores}, \binits{M.}},
\bauthor{\bsnm{Costa}, \binits{A.B.}},
\bauthor{\bsnm{Aphinyanaphongs}, \binits{Y.}},
\bauthor{\bsnm{Cho}, \binits{K.}},
\bauthor{\bsnm{Oermann}, \binits{E.K.}}:
\batitle{Health system-scale language models are all-purpose prediction engines}.
\bjtitle{Nature}
\bvolume{619}(\bissue{7969}),
\bfpage{357}--\blpage{362}
(\byear{2023})
\doiurl{10.1038/s41586-023-06160-y}
\end{barticle}
\endbibitem

\bibitem[\protect\citeauthoryear{Takita et~al.}{2024}]{takita2024diagnostic}
\begin{botherref}
\oauthor{\bsnm{Takita}, \binits{H.}},
\oauthor{\bsnm{Walston}, \binits{S.L.}},
\oauthor{\bsnm{Tatekawa}, \binits{H.}},
\oauthor{\bsnm{Saito}, \binits{K.}},
\oauthor{\bsnm{Tsujimoto}, \binits{Y.}},
\oauthor{\bsnm{Miki}, \binits{Y.}},
\oauthor{\bsnm{Ueda}, \binits{D.}}:
Diagnostic performance of generative {AI} and physicians: A systematic review and meta-analysis.
Cold Spring Harbor Laboratory Press
(2024).
\doiurl{10.1101/2024.01.20.24301563}
\end{botherref}
\endbibitem

\bibitem[\protect\citeauthoryear{Chakravorti}{2024, May 3}]{chakravortiAITrustProblem2024}
\begin{botherref}
\oauthor{\bsnm{Chakravorti}, \binits{B.}}:
{AI}'s trust problem.
Harvard Business Review
(2024, May 3).
\url{https://hbr.org/2024/05/ais-trust-problem}
\end{botherref}
\endbibitem

\bibitem[\protect\citeauthoryear{Lenat and Marcus}{2023}]{lenat2023getting}
\begin{botherref}
\oauthor{\bsnm{Lenat}, \binits{D.}},
\oauthor{\bsnm{Marcus}, \binits{G.}}:
Getting from generative AI to trustworthy AI: What LLMs might learn from Cyc.
arXiv
(2023).
\url{https://arxiv.org/abs/2308.04445}
\end{botherref}
\endbibitem

\bibitem[\protect\citeauthoryear{Cowls et~al.}{2023}]{Cowls-ClimateAIGambit-2023}
\begin{barticle}
\bauthor{\bsnm{Cowls}, \binits{J.}},
\bauthor{\bsnm{Tsamados}, \binits{A.}},
\bauthor{\bsnm{Taddeo}, \binits{M.}},
\bauthor{\bsnm{Floridi}, \binits{L.}}:
\batitle{{The AI gambit: Leveraging artificial intelligence to combat climate change—opportunities, challenges, and recommendations}}.
\bjtitle{AI \& SOCIETY}
\bvolume{38}(\bissue{1}),
\bfpage{283}--\blpage{307}
(\byear{2023})
\doiurl{10.1007/s00146-021-01294-x}
\end{barticle}
\endbibitem

\bibitem[\protect\citeauthoryear{Woolley et~al.}{2010}]{Woolley:kh2010}
\begin{barticle}
\bauthor{\bsnm{Woolley}, \binits{A.W.}},
\bauthor{\bsnm{Chabris}, \binits{C.F.}},
\bauthor{\bsnm{Pentland}, \binits{A.}},
\bauthor{\bsnm{Hashmi}, \binits{N.}},
\bauthor{\bsnm{Malone}, \binits{T.W.}}:
\batitle{Evidence for a collective intelligence factor in the performance of human groups}.
\bjtitle{Science}
\bvolume{330}(\bissue{6004}),
\bfpage{686}--\blpage{688}
(\byear{2010})
\doiurl{10.1126/science.1193147}
\end{barticle}
\endbibitem

\bibitem[\protect\citeauthoryear{Woolley and Gupta}{2024}]{Woolley-2024}
\begin{barticle}
\bauthor{\bsnm{Woolley}, \binits{A.W.}},
\bauthor{\bsnm{Gupta}, \binits{P.}}:
\batitle{Understanding collective intelligence: Investigating the role of collective memory, attention, and reasoning processes}.
\bjtitle{Perspectives on Psychological Science}
\bvolume{19}(\bissue{2}),
\bfpage{344}--\blpage{354}
(\byear{2024})
\doiurl{10.1177/17456916231191534}
\end{barticle}
\endbibitem

\bibitem[\protect\citeauthoryear{Hasan et~al.}{2024}]{hasan2024boosting}
\begin{barticle}
\bauthor{\bsnm{Hasan}, \binits{E.}},
\bauthor{\bsnm{Duhaime}, \binits{E.}},
\bauthor{\bsnm{Trueblood}, \binits{J.S.}}:
\batitle{Boosting wisdom of the crowd for medical image annotation using training performance and task features}.
\bjtitle{Cognitive Research: Principles and Implications}
\bvolume{9}(\bissue{1}),
\bfpage{31}
(\byear{2024})
\doiurl{10.1186/s41235-024-00558-6}
\end{barticle}
\endbibitem

\bibitem[\protect\citeauthoryear{Hautz et~al.}{2015}]{hautz2015diagnostic}
\begin{barticle}
\bauthor{\bsnm{Hautz}, \binits{W.E.}},
\bauthor{\bsnm{K{\"a}mmer}, \binits{J.E.}},
\bauthor{\bsnm{Schauber}, \binits{S.K.}},
\bauthor{\bsnm{Spies}, \binits{C.D.}},
\bauthor{\bsnm{Gaissmaier}, \binits{W.}}:
\batitle{Diagnostic performance by medical students working individually or in teams}.
\bjtitle{JAMA}
\bvolume{313}(\bissue{3}),
\bfpage{303}--\blpage{304}
(\byear{2015})
\doiurl{10.1001/jama.2014.15770}
\end{barticle}
\endbibitem

\bibitem[\protect\citeauthoryear{Kattan et~al.}{2016}]{kattan2016wisdom}
\begin{barticle}
\bauthor{\bsnm{Kattan}, \binits{M.W.}},
\bauthor{\bsnm{O’Rourke}, \binits{C.}},
\bauthor{\bsnm{Yu}, \binits{C.}},
\bauthor{\bsnm{Chagin}, \binits{K.}}:
\batitle{The wisdom of crowds of doctors: Their average predictions outperform their individual ones}.
\bjtitle{Medical Decision Making}
\bvolume{36}(\bissue{4}),
\bfpage{536}--\blpage{540}
(\byear{2016})
\doiurl{10.1177/0272989X15581615}
\end{barticle}
\endbibitem

\bibitem[\protect\citeauthoryear{Kurvers et~al.}{2016}]{kurvers2016boosting}
\begin{barticle}
\bauthor{\bsnm{Kurvers}, \binits{R.H.}},
\bauthor{\bsnm{Herzog}, \binits{S.M.}},
\bauthor{\bsnm{Hertwig}, \binits{R.}},
\bauthor{\bsnm{Krause}, \binits{J.}},
\bauthor{\bsnm{Carney}, \binits{P.A.}},
\bauthor{\bsnm{Bogart}, \binits{A.}},
\bauthor{\bsnm{Argenziano}, \binits{G.}},
\bauthor{\bsnm{Zalaudek}, \binits{I.}},
\bauthor{\bsnm{Wolf}, \binits{M.}}:
\batitle{Boosting medical diagnostics by pooling independent judgments}.
\bjtitle{Proceedings of the National Academy of Sciences}
\bvolume{113}(\bissue{31}),
\bfpage{8777}--\blpage{8782}
(\byear{2016})
\doiurl{10.1073/pnas.1601827113}
\end{barticle}
\endbibitem

\bibitem[\protect\citeauthoryear{K{\"a}mmer et~al.}{2017}]{kammer2017potential}
\begin{barticle}
\bauthor{\bsnm{K{\"a}mmer}, \binits{J.E.}},
\bauthor{\bsnm{Hautz}, \binits{W.E.}},
\bauthor{\bsnm{Herzog}, \binits{S.M.}},
\bauthor{\bsnm{Kunina-Habenicht}, \binits{O.}},
\bauthor{\bsnm{Kurvers}, \binits{R.H.}}:
\batitle{The potential of collective intelligence in emergency medicine: Pooling medical students’ independent decisions improves diagnostic performance}.
\bjtitle{Medical Decision Making}
\bvolume{37}(\bissue{6}),
\bfpage{715}--\blpage{724}
(\byear{2017})
\doiurl{10.1177/0272989X17696998}
\end{barticle}
\endbibitem

\bibitem[\protect\citeauthoryear{Kurvers et~al.}{2018}]{kurvers2018combining}
\begin{barticle}
\bauthor{\bsnm{Kurvers}, \binits{R.H.}},
\bauthor{\bsnm{De~Zoete}, \binits{A.}},
\bauthor{\bsnm{Bachman}, \binits{S.L.}},
\bauthor{\bsnm{Algra}, \binits{P.R.}},
\bauthor{\bsnm{Ostelo}, \binits{R.}}:
\batitle{Combining independent decisions increases diagnostic accuracy of reading lumbosacral radiographs and magnetic resonance imaging}.
\bjtitle{PloS One}
\bvolume{13}(\bissue{4}),
\bfpage{0194128}
(\byear{2018})
\doiurl{10.1371/journal.pone.0194128}
\end{barticle}
\endbibitem

\bibitem[\protect\citeauthoryear{Blanchard et~al.}{2024}]{Blanchard2024}
\begin{barticle}
\bauthor{\bsnm{Blanchard}, \binits{M.D.}},
\bauthor{\bsnm{Herzog}, \binits{S.M.}},
\bauthor{\bsnm{K{\"a}mmer}, \binits{J.E.}},
\bauthor{\bsnm{Z{\"o}ller}, \binits{N.}},
\bauthor{\bsnm{Kostopoulou}, \binits{O.}},
\bauthor{\bsnm{Kurvers}, \binits{R.H.J.M.}}:
\batitle{Collective intelligence increases diagnostic accuracy in a general practice setting}.
\bjtitle{Medical Decision Making}
\bvolume{44}(\bissue{4}),
\bfpage{451}--\blpage{462}
(\byear{2024})
\doiurl{10.1177/0272989X241241001}
\end{barticle}
\endbibitem

\bibitem[\protect\citeauthoryear{Barnett et~al.}{2019}]{barnett2019comparative}
\begin{barticle}
\bauthor{\bsnm{Barnett}, \binits{M.L.}},
\bauthor{\bsnm{Boddupalli}, \binits{D.}},
\bauthor{\bsnm{Nundy}, \binits{S.}},
\bauthor{\bsnm{Bates}, \binits{D.W.}}:
\batitle{Comparative accuracy of diagnosis by collective intelligence of multiple physicians vs individual physicians}.
\bjtitle{JAMA Network Open}
\bvolume{2}(\bissue{3}),
\bfpage{190096}
(\byear{2019})
\doiurl{10.1001/jamanetworkopen.2019.0096}
\end{barticle}
\endbibitem

\bibitem[\protect\citeauthoryear{Kurvers et~al.}{2023}]{kurvers2023}
\begin{barticle}
\bauthor{\bsnm{Kurvers}, \binits{R.H.J.M.}},
\bauthor{\bsnm{Nuzzolese}, \binits{A.G.}},
\bauthor{\bsnm{Russo}, \binits{A.}},
\bauthor{\bsnm{Barabucci}, \binits{G.}},
\bauthor{\bsnm{Herzog}, \binits{S.M.}},
\bauthor{\bsnm{Trianni}, \binits{V.}}:
\batitle{Automating hybrid collective intelligence in open-ended medical diagnostics}.
\bjtitle{Proceedings of the National Academy of Sciences}
\bvolume{120}(\bissue{34}),
\bfpage{2221473120}
(\byear{2023})
\doiurl{10.1073/pnas.2221473120}
\end{barticle}
\endbibitem

\bibitem[\protect\citeauthoryear{Kuncheva}{}]{kuncheva2014combining}
\begin{botherref}
\oauthor{\bsnm{Kuncheva}, \binits{L.I.}}:
Combining Pattern Classifiers: Methods and Algorithms,
2. ed edn.
Wiley.
\doiurl{10.1002/9781118914564}
\end{botherref}
\endbibitem

\bibitem[\protect\citeauthoryear{Mienye and Sun}{2022}]{mienye2022survey}
\begin{barticle}
\bauthor{\bsnm{Mienye}, \binits{I.D.}},
\bauthor{\bsnm{Sun}, \binits{Y.}}:
\batitle{A survey of ensemble learning: Concepts, algorithms, applications, and prospects}.
\bjtitle{IEEE Access}
\bvolume{10},
\bfpage{99129}--\blpage{99149}
(\byear{2022})
\doiurl{10.1109/ACCESS.2022.3207287}
\end{barticle}
\endbibitem

\bibitem[\protect\citeauthoryear{Jiang et~al.}{2023}]{jiang2023llm}
\begin{botherref}
\oauthor{\bsnm{Jiang}, \binits{D.}},
\oauthor{\bsnm{Ren}, \binits{X.}},
\oauthor{\bsnm{Lin}, \binits{B.Y.}}:
{LLM-Blender}: Ensembling large language models with pairwise ranking and generative fusion
(2023).
\url{https://arxiv.org/abs/2306.02561}
\end{botherref}
\endbibitem

\bibitem[\protect\citeauthoryear{Jiang et~al.}{2024}]{jiang2024mixtral}
\begin{botherref}
\oauthor{\bsnm{Jiang}, \binits{A.Q.}},
\oauthor{\bsnm{Sablayrolles}, \binits{A.}},
\oauthor{\bsnm{Roux}, \binits{A.}},
\oauthor{\bsnm{Mensch}, \binits{A.}},
\oauthor{\bsnm{Savary}, \binits{B.}},
\oauthor{\bsnm{Bamford}, \binits{C.}},
\oauthor{\bsnm{Chaplot}, \binits{D.S.}},
\oauthor{\bsnm{Casas}, \binits{D.}},
\oauthor{\bsnm{Hanna}, \binits{E.B.}},
\oauthor{\bsnm{Bressand}, \binits{F.}},
\oauthor{\bsnm{Lengyel}, \binits{G.}},
\oauthor{\bsnm{Bour}, \binits{G.}},
\oauthor{\bsnm{Lample}, \binits{G.}},
\oauthor{\bsnm{Lavaud}, \binits{L.R.}},
\oauthor{\bsnm{Saulnier}, \binits{L.}},
\oauthor{\bsnm{Lachaux}, \binits{M.-A.}},
\oauthor{\bsnm{Stock}, \binits{P.}},
\oauthor{\bsnm{Subramanian}, \binits{S.}},
\oauthor{\bsnm{Yang}, \binits{S.}},
\oauthor{\bsnm{Antoniak}, \binits{S.}},
\oauthor{\bsnm{Scao}, \binits{T.L.}},
\oauthor{\bsnm{Gervet}, \binits{T.}},
\oauthor{\bsnm{Lavril}, \binits{T.}},
\oauthor{\bsnm{Wang}, \binits{T.}},
\oauthor{\bsnm{Lacroix}, \binits{T.}},
\oauthor{\bsnm{Sayed}, \binits{W.E.}}:
Mixtral of experts.
arXiv
(2024).
\url{https://arxiv.org/abs/2401.04088}
\end{botherref}
\endbibitem

\bibitem[\protect\citeauthoryear{Yang et~al.}{2023}]{yang2023one}
\begin{botherref}
\oauthor{\bsnm{Yang}, \binits{H.}},
\oauthor{\bsnm{Li}, \binits{M.}},
\oauthor{\bsnm{Xiao}, \binits{Y.}},
\oauthor{\bsnm{Zhou}, \binits{H.}},
\oauthor{\bsnm{Zhang}, \binits{R.}},
\oauthor{\bsnm{Fang}, \binits{Q.}}:
One {LLM} is not enough: Harnessing the power of ensemble learning for medical question answering.
medRxiv
(2023).
\url{https://www.medrxiv.org/content/10.1101/2023.12.21.23300380v1}
\end{botherref}
\endbibitem

\bibitem[\protect\citeauthoryear{Barabucci et~al.}{2024}]{barabucci2024combining}
\begin{botherref}
\oauthor{\bsnm{Barabucci}, \binits{G.}},
\oauthor{\bsnm{Shia}, \binits{V.}},
\oauthor{\bsnm{Chu}, \binits{E.}},
\oauthor{\bsnm{Harack}, \binits{B.}},
\oauthor{\bsnm{Fu}, \binits{N.}}:
Combining insights from multiple large language models improves diagnostic accuracy.
arXiv
(2024).
\url{https://arxiv.org/abs/2402.08806}
\end{botherref}
\endbibitem

\bibitem[\protect\citeauthoryear{Dogan and Birant}{2019}]{Dogan2019}
\begin{bchapter}
\bauthor{\bsnm{Dogan}, \binits{A.}},
\bauthor{\bsnm{Birant}, \binits{D.}}:
\bctitle{A weighted majority voting ensemble approach for classification}.
In: \bbtitle{2019 4th International Conference on Computer Science and Engineering (UBMK)}.
\bpublisher{IEEE},
\blocation{Samsun, Turkey}
(\byear{2019}).
\doiurl{10.1109/ubmk.2019.8907028}
\end{bchapter}
\endbibitem

\bibitem[\protect\citeauthoryear{Donnelly}{2006}]{donnelly2006snomed}
\begin{barticle}
\bauthor{\bsnm{Donnelly}, \binits{K.}}:
\batitle{{SNOMED-CT}: The advanced terminology and coding system for {eHealth}}.
\bjtitle{Studies in Health Technology and Informatics}
\bvolume{121},
\bfpage{279}--\blpage{290}
(\byear{2006})
\end{barticle}
\endbibitem

\bibitem[\protect\citeauthoryear{Voorhees}{1999}]{voorhees1999trec}
\begin{bchapter}
\bauthor{\bsnm{Voorhees}, \binits{E.M.}}:
\bctitle{The {TREC-8} question answering track report}.
In: \beditor{\bsnm{Voorhees}, \binits{E.M.}},
\beditor{\bsnm{Harman}, \binits{D.K.}} (eds.)
\bbtitle{Proceedings of The Eighth Text REtrieval Conference, TREC 1999}.
\bsertitle{{NIST} Special Publication},
vol. \bseriesno{500-246},
pp. \bfpage{77}--\blpage{82}.
\bpublisher{National Institute of Standards and Technology {(NIST)}},
\blocation{Gaithersburg, MA}
(\byear{1999}).
\burl{https://trec.nist.gov/pubs/trec8/t8{\_}proceedings.html}
\end{bchapter}
\endbibitem

\bibitem[\protect\citeauthoryear{Ladha}{1992}]{ladha1992condorcet}
\begin{barticle}
\bauthor{\bsnm{Ladha}, \binits{K.K.}}:
\batitle{The {C}ondorcet jury theorem, free speech, and correlated votes}.
\bjtitle{American Journal of Political Science}
\bvolume{36}(\bissue{3}),
\bfpage{617}--\blpage{634}
(\byear{1992})
\doiurl{10.2307/2111584}
\end{barticle}
\endbibitem

\bibitem[\protect\citeauthoryear{Grofman et~al.}{1983}]{grofman1983thirteen}
\begin{barticle}
\bauthor{\bsnm{Grofman}, \binits{B.}},
\bauthor{\bsnm{Owen}, \binits{G.}},
\bauthor{\bsnm{Feld}, \binits{S.L.}}:
\batitle{Thirteen theorems in search of the truth}.
\bjtitle{Theory and Decision}
\bvolume{15}(\bissue{3}),
\bfpage{261}--\blpage{278}
(\byear{1983})
\doiurl{10.1007/BF00125672}
\end{barticle}
\endbibitem

\bibitem[\protect\citeauthoryear{Tumer and Ghosh}{1996}]{tumer1996error}
\begin{barticle}
\bauthor{\bsnm{Tumer}, \binits{K.}},
\bauthor{\bsnm{Ghosh}, \binits{J.}}:
\batitle{Error correlation and error reduction in ensemble classifiers}.
\bjtitle{Connection Science}
\bvolume{8}(\bissue{3-4}),
\bfpage{385}--\blpage{404}
(\byear{1996})
\doiurl{10.1080/095400996116839}
\end{barticle}
\endbibitem

\bibitem[\protect\citeauthoryear{Marshall et~al.}{2019}]{marshall2019quorums}
\begin{barticle}
\bauthor{\bsnm{Marshall}, \binits{J.A.}},
\bauthor{\bsnm{Kurvers}, \binits{R.H.}},
\bauthor{\bsnm{Krause}, \binits{J.}},
\bauthor{\bsnm{Wolf}, \binits{M.}}:
\batitle{Quorums enable optimal pooling of independent judgements in biological systems}.
\bjtitle{eLife}
\bvolume{8},
\bfpage{40368}
(\byear{2019})
\doiurl{10.7554/eLife.40368}
\end{barticle}
\endbibitem

\bibitem[\protect\citeauthoryear{Steyvers et~al.}{2022}]{steyvers2022bayesian}
\begin{barticle}
\bauthor{\bsnm{Steyvers}, \binits{M.}},
\bauthor{\bsnm{Tejeda}, \binits{H.}},
\bauthor{\bsnm{Kerrigan}, \binits{G.}},
\bauthor{\bsnm{Smyth}, \binits{P.}}:
\batitle{Bayesian modeling of human--{AI} complementarity}.
\bjtitle{Proceedings of the National Academy of Sciences}
\bvolume{119}(\bissue{11}),
\bfpage{2111547119}
(\byear{2022})
\doiurl{10.1073/pnas.211154711}
\end{barticle}
\endbibitem

\bibitem[\protect\citeauthoryear{Pescetelli}{2021}]{Pescetelli2021}
\begin{barticle}
\bauthor{\bsnm{Pescetelli}, \binits{N.}}:
\batitle{A brief taxonomy of hybrid intelligence}.
\bjtitle{Forecasting}
\bvolume{3}(\bissue{3}),
\bfpage{633}--\blpage{643}
(\byear{2021})
\doiurl{10.3390/forecast3030039}
\end{barticle}
\endbibitem

\bibitem[\protect\citeauthoryear{Peeters et~al.}{2020}]{Peeters2020}
\begin{barticle}
\bauthor{\bsnm{Peeters}, \binits{M.M.M.}},
\bauthor{\bsnm{Diggelen}, \binits{J.}},
\bauthor{\bsnm{Bosch}, \binits{K.}},
\bauthor{\bsnm{Bronkhorst}, \binits{A.}},
\bauthor{\bsnm{Neerincx}, \binits{M.A.}},
\bauthor{\bsnm{Schraagen}, \binits{J.M.}},
\bauthor{\bsnm{Raaijmakers}, \binits{S.}}:
\batitle{Hybrid collective intelligence in a human–{AI} society}.
\bjtitle{AI \& Society}
\bvolume{36}(\bissue{1}),
\bfpage{217}--\blpage{238}
(\byear{2020})
\doiurl{10.1007/s00146-020-01005-y}
\end{barticle}
\endbibitem

\bibitem[\protect\citeauthoryear{Steyvers and Kumar}{2023}]{Steyvers2023}
\begin{barticle}
\bauthor{\bsnm{Steyvers}, \binits{M.}},
\bauthor{\bsnm{Kumar}, \binits{A.}}:
\batitle{Three challenges for {AI}-assisted decision-making}.
\bjtitle{Perspectives on Psychological Science}
(\byear{2023})
\doiurl{10.1177/17456916231181102}
\end{barticle}
\endbibitem

\bibitem[\protect\citeauthoryear{Benjamin et~al.}{2023}]{Benjamin2023}
\begin{barticle}
\bauthor{\bsnm{Benjamin}, \binits{D.M.}},
\bauthor{\bsnm{Morstatter}, \binits{F.}},
\bauthor{\bsnm{Abbas}, \binits{A.E.}},
\bauthor{\bsnm{Abeliuk}, \binits{A.}},
\bauthor{\bsnm{Atanasov}, \binits{P.}},
\bauthor{\bsnm{Bennett}, \binits{S.}},
\bauthor{\bsnm{Beger}, \binits{A.}},
\bauthor{\bsnm{Birari}, \binits{S.}},
\bauthor{\bsnm{Budescu}, \binits{D.V.}},
\bauthor{\bsnm{Catasta}, \binits{M.}},
\bauthor{\bsnm{Ferrara}, \binits{E.}},
\bauthor{\bsnm{Haravitch}, \binits{L.}},
\bauthor{\bsnm{Himmelstein}, \binits{M.}},
\bauthor{\bsnm{Hossain}, \binits{K.T.}},
\bauthor{\bsnm{Huang}, \binits{Y.}},
\bauthor{\bsnm{Jin}, \binits{W.}},
\bauthor{\bsnm{Joseph}, \binits{R.}},
\bauthor{\bsnm{Leskovec}, \binits{J.}},
\bauthor{\bsnm{Matsui}, \binits{A.}},
\bauthor{\bsnm{Mirtaheri}, \binits{M.}},
\bauthor{\bsnm{Ren}, \binits{X.}},
\bauthor{\bsnm{Satyukov}, \binits{G.}},
\bauthor{\bsnm{Sethi}, \binits{R.}},
\bauthor{\bsnm{Singh}, \binits{A.}},
\bauthor{\bsnm{Sosic}, \binits{R.}},
\bauthor{\bsnm{Steyvers}, \binits{M.}},
\bauthor{\bsnm{Szekely}, \binits{P.A.}},
\bauthor{\bsnm{Ward}, \binits{M.D.}},
\bauthor{\bsnm{Galstyan}, \binits{A.}}:
\batitle{Hybrid forecasting of geopolitical events}.
\bjtitle{AI Magazine}
\bvolume{44}(\bissue{1}),
\bfpage{112}--\blpage{128}
(\byear{2023})
\doiurl{10.1002/aaai.12085}
\end{barticle}
\endbibitem

\bibitem[\protect\citeauthoryear{Peabody et~al.}{2004}]{peabody2004measuring}
\begin{barticle}
\bauthor{\bsnm{Peabody}, \binits{J.W.}},
\bauthor{\bsnm{Luck}, \binits{J.}},
\bauthor{\bsnm{Glassman}, \binits{P.}},
\bauthor{\bsnm{Jain}, \binits{S.}},
\bauthor{\bsnm{Hansen}, \binits{J.}},
\bauthor{\bsnm{Spell}, \binits{M.}},
\bauthor{\bsnm{Lee}, \binits{M.}}:
\batitle{Measuring the quality of physician practice by using clinical vignettes: A prospective validation study}.
\bjtitle{Annals of Internal Medicine}
\bvolume{141}(\bissue{10}),
\bfpage{771}--\blpage{780}
(\byear{2004})
\doiurl{10.7326/0003-4819-141-10-200411160-00008}
\end{barticle}
\endbibitem

\bibitem[\protect\citeauthoryear{Yabroff et~al.}{2020}]{Yabroff2020}
\begin{barticle}
\bauthor{\bsnm{Yabroff}, \binits{K.R.}},
\bauthor{\bsnm{Reeder-Hayes}, \binits{K.}},
\bauthor{\bsnm{Zhao}, \binits{J.}},
\bauthor{\bsnm{Halpern}, \binits{M.T.}},
\bauthor{\bsnm{Lopez}, \binits{A.M.}},
\bauthor{\bsnm{Bernal-Mizrachi}, \binits{L.}},
\bauthor{\bsnm{Collier}, \binits{A.B.}},
\bauthor{\bsnm{Neuner}, \binits{J.}},
\bauthor{\bsnm{Phillips}, \binits{J.}},
\bauthor{\bsnm{Blackstock}, \binits{W.}},
\bauthor{\bsnm{Patel}, \binits{M.}}:
\batitle{Health insurance coverage disruptions and cancer care and outcomes: Systematic review of published research}.
\bjtitle{JNCI: Journal of the National Cancer Institute}
\bvolume{112}(\bissue{7}),
\bfpage{671}--\blpage{687}
(\byear{2020})
\doiurl{10.1093/jnci/djaa048}
\end{barticle}
\endbibitem

\bibitem[\protect\citeauthoryear{Hooker}{2021}]{hookerMovingAlgorithmicBias2021}
\begin{botherref}
\oauthor{\bsnm{Hooker}, \binits{S.}}:
Moving beyond “algorithmic bias is a data problem”.
Patterns
\textbf{2}(4)
(2021)
\doiurl{10.1016/j.patter.2021.100241}
{\href{https://arxiv.org/abs/33982031}{{33982031}}}
\end{botherref}
\endbibitem

\bibitem[\protect\citeauthoryear{{van Giffen} et~al.}{2022}]{vangiffenOvercomingPitfallsPerils2022}
\begin{barticle}
\bauthor{\bsnm{{van Giffen}}, \binits{B.}},
\bauthor{\bsnm{Herhausen}, \binits{D.}},
\bauthor{\bsnm{Fahse}, \binits{T.}}:
\batitle{Overcoming the pitfalls and perils of algorithms: {{A}} classification of machine learning biases and mitigation methods}.
\bjtitle{Journal of Business Research}
\bvolume{144},
\bfpage{93}--\blpage{106}
(\byear{2022})
\doiurl{10.1016/j.jbusres.2022.01.076}
\end{barticle}
\endbibitem

\bibitem[\protect\citeauthoryear{Wachter et~al.}{2021}]{wachterBiasPreservationMachine2021}
\begin{barticle}
\bauthor{\bsnm{Wachter}, \binits{S.}},
\bauthor{\bsnm{Mittelstadt}, \binits{B.}},
\bauthor{\bsnm{Russell}, \binits{C.}}:
\batitle{Bias preservation in machine learning: {{The}} legality of fairness metrics under {{EU}} non-discrimination law}.
\bjtitle{West Virginia Law Review}
\bvolume{123},
\bfpage{735}--\blpage{790}
(\byear{2021})
\doiurl{10.2139/ssrn.3792772}
\end{barticle}
\endbibitem

\bibitem[\protect\citeauthoryear{Weidinger et~al.}{2021}]{weidingerEthicalSocialRisks2021}
\begin{botherref}
\oauthor{\bsnm{Weidinger}, \binits{L.}},
\oauthor{\bsnm{Mellor}, \binits{J.}},
\oauthor{\bsnm{Rauh}, \binits{M.}},
\oauthor{\bsnm{Griffin}, \binits{C.}},
\oauthor{\bsnm{Uesato}, \binits{J.}},
\oauthor{\bsnm{Huang}, \binits{P.-S.}},
\oauthor{\bsnm{Cheng}, \binits{M.}},
\oauthor{\bsnm{Glaese}, \binits{M.}},
\oauthor{\bsnm{Balle}, \binits{B.}},
\oauthor{\bsnm{Kasirzadeh}, \binits{A.}},
\oauthor{\bsnm{Kenton}, \binits{Z.}},
\oauthor{\bsnm{Brown}, \binits{S.}},
\oauthor{\bsnm{Hawkins}, \binits{W.}},
\oauthor{\bsnm{Stepleton}, \binits{T.}},
\oauthor{\bsnm{Biles}, \binits{C.}},
\oauthor{\bsnm{Birhane}, \binits{A.}},
\oauthor{\bsnm{Haas}, \binits{J.}},
\oauthor{\bsnm{Rimell}, \binits{L.}},
\oauthor{\bsnm{Hendricks}, \binits{L.A.}},
\oauthor{\bsnm{Isaac}, \binits{W.}},
\oauthor{\bsnm{Legassick}, \binits{S.}},
\oauthor{\bsnm{Irving}, \binits{G.}},
\oauthor{\bsnm{Gabriel}, \binits{I.}}:
Ethical and social risks of harm from language models.
arXiv
(2021).
\url{http://arxiv.org/abs/2112.04359}
\end{botherref}
\endbibitem

\bibitem[\protect\citeauthoryear{Liang et~al.}{2023}]{liangHolisticEvaluationLanguage2023}
\begin{botherref}
\oauthor{\bsnm{Liang}, \binits{P.}},
\oauthor{\bsnm{Bommasani}, \binits{R.}},
\oauthor{\bsnm{Lee}, \binits{T.}},
\oauthor{\bsnm{Tsipras}, \binits{D.}},
\oauthor{\bsnm{Soylu}, \binits{D.}},
\oauthor{\bsnm{Yasunaga}, \binits{M.}},
\oauthor{\bsnm{Zhang}, \binits{Y.}},
\oauthor{\bsnm{Narayanan}, \binits{D.}},
\oauthor{\bsnm{Wu}, \binits{Y.}},
\oauthor{\bsnm{Kumar}, \binits{A.}},
\oauthor{\bsnm{Newman}, \binits{B.}},
\oauthor{\bsnm{Yuan}, \binits{B.}},
\oauthor{\bsnm{Yan}, \binits{B.}},
\oauthor{\bsnm{Zhang}, \binits{C.}},
\oauthor{\bsnm{Cosgrove}, \binits{C.}},
\oauthor{\bsnm{Manning}, \binits{C.D.}},
\oauthor{\bsnm{Ré}, \binits{C.}},
\oauthor{\bsnm{Acosta-Navas}, \binits{D.}},
\oauthor{\bsnm{Hudson}, \binits{D.A.}},
\oauthor{\bsnm{Zelikman}, \binits{E.}},
\oauthor{\bsnm{Durmus}, \binits{E.}},
\oauthor{\bsnm{Ladhak}, \binits{F.}},
\oauthor{\bsnm{Rong}, \binits{F.}},
\oauthor{\bsnm{Ren}, \binits{H.}},
\oauthor{\bsnm{Yao}, \binits{H.}},
\oauthor{\bsnm{Wang}, \binits{J.}},
\oauthor{\bsnm{Santhanam}, \binits{K.}},
\oauthor{\bsnm{Orr}, \binits{L.}},
\oauthor{\bsnm{Zheng}, \binits{L.}},
\oauthor{\bsnm{Yuksekgonul}, \binits{M.}},
\oauthor{\bsnm{Suzgun}, \binits{M.}},
\oauthor{\bsnm{Kim}, \binits{N.}},
\oauthor{\bsnm{Guha}, \binits{N.}},
\oauthor{\bsnm{Chatterji}, \binits{N.}},
\oauthor{\bsnm{Khattab}, \binits{O.}},
\oauthor{\bsnm{Henderson}, \binits{P.}},
\oauthor{\bsnm{Huang}, \binits{Q.}},
\oauthor{\bsnm{Chi}, \binits{R.}},
\oauthor{\bsnm{Xie}, \binits{S.M.}},
\oauthor{\bsnm{Santurkar}, \binits{S.}},
\oauthor{\bsnm{Ganguli}, \binits{S.}},
\oauthor{\bsnm{Hashimoto}, \binits{T.}},
\oauthor{\bsnm{Icard}, \binits{T.}},
\oauthor{\bsnm{Zhang}, \binits{T.}},
\oauthor{\bsnm{Chaudhary}, \binits{V.}},
\oauthor{\bsnm{Wang}, \binits{W.}},
\oauthor{\bsnm{Li}, \binits{X.}},
\oauthor{\bsnm{Mai}, \binits{Y.}},
\oauthor{\bsnm{Zhang}, \binits{Y.}},
\oauthor{\bsnm{Koreeda}, \binits{Y.}}:
Holistic evaluation of language models.
arXiv
(2023).
\url{https://arxiv.org/abs/2211.09110}
\end{botherref}
\endbibitem

\bibitem[\protect\citeauthoryear{Paulus and Kent}{2020}]{paulus2020predictably}
\begin{barticle}
\bauthor{\bsnm{Paulus}, \binits{J.K.}},
\bauthor{\bsnm{Kent}, \binits{D.M.}}:
\batitle{Predictably unequal: Understanding and addressing concerns that algorithmic clinical prediction may increase health disparities}.
\bjtitle{npj Digital Medicine}
\bvolume{3}(\bissue{1}),
\bfpage{99}
(\byear{2020})
\doiurl{10.1038/s41746-020-0304-9}
\end{barticle}
\endbibitem

\bibitem[\protect\citeauthoryear{Groh et~al.}{2024}]{Groh2024}
\begin{barticle}
\bauthor{\bsnm{Groh}, \binits{M.}},
\bauthor{\bsnm{Badri}, \binits{O.}},
\bauthor{\bsnm{Daneshjou}, \binits{R.}},
\bauthor{\bsnm{Koochek}, \binits{A.}},
\bauthor{\bsnm{Harris}, \binits{C.}},
\bauthor{\bsnm{Soenksen}, \binits{L.R.}},
\bauthor{\bsnm{Doraiswamy}, \binits{P.M.}},
\bauthor{\bsnm{Picard}, \binits{R.}}:
\batitle{Deep learning-aided decision support for diagnosis of skin disease across skin tones}.
\bjtitle{Nature Medicine}
\bvolume{30}(\bissue{2}),
\bfpage{573}--\blpage{583}
(\byear{2024})
\doiurl{10.1038/s41591-023-02728-3}
\end{barticle}
\endbibitem

\bibitem[\protect\citeauthoryear{Birhane et~al.}{2022}]{Birhane2022power}
\begin{bchapter}
\bauthor{\bsnm{Birhane}, \binits{A.}},
\bauthor{\bsnm{Isaac}, \binits{W.}},
\bauthor{\bsnm{Prabhakaran}, \binits{V.}},
\bauthor{\bsnm{Diaz}, \binits{M.}},
\bauthor{\bsnm{Elish}, \binits{M.C.}},
\bauthor{\bsnm{Gabriel}, \binits{I.}},
\bauthor{\bsnm{Mohamed}, \binits{S.}}:
\bctitle{Power to the people? {O}pportunities and challenges for participatory {AI}}.
In: \bbtitle{Proceedings of the 2nd ACM Conference on Equity and Access in Algorithms, Mechanisms, and Optimization}.
\bsertitle{EAAMO '22}.
\bpublisher{Association for Computing Machinery},
\blocation{New York, NY}
(\byear{2022}).
\doiurl{10.1145/3551624.3555290}
\end{bchapter}
\endbibitem

\bibitem[\protect\citeauthoryear{Carusi et~al.}{}]{carusiMedicalArtificialIntelligence2023}
\begin{botherref}
\oauthor{\bsnm{Carusi}, \binits{A.}},
\oauthor{\bsnm{Winter}, \binits{P.D.}},
\oauthor{\bsnm{Armstrong}, \binits{I.}},
\oauthor{\bsnm{Ciravegna}, \binits{F.}},
\oauthor{\bsnm{Kiely}, \binits{D.G.}},
\oauthor{\bsnm{Lawrie}, \binits{A.}},
\oauthor{\bsnm{Lu}, \binits{H.}},
\oauthor{\bsnm{Sabroe}, \binits{I.}},
\oauthor{\bsnm{Swift}, \binits{A.}}:
Medical artificial intelligence is as much social as it is technological.
Nature Machine Intelligence
\textbf{5}(2),
98--100
\doiurl{10.1038/s42256-022-00603-3}
\end{botherref}
\endbibitem

\bibitem[\protect\citeauthoryear{Delgado et~al.}{2023}]{Delgado2023participatory}
\begin{bchapter}
\bauthor{\bsnm{Delgado}, \binits{F.}},
\bauthor{\bsnm{Yang}, \binits{S.}},
\bauthor{\bsnm{Madaio}, \binits{M.}},
\bauthor{\bsnm{Yang}, \binits{Q.}}:
\bctitle{The participatory turn in {AI} design: Theoretical foundations and the current state of practice}.
In: \bbtitle{Proceedings of the 3rd ACM Conference on Equity and Access in Algorithms, Mechanisms, and Optimization}.
\bsertitle{EAAMO '23}.
\bpublisher{Association for Computing Machinery},
\blocation{New York, NY}
(\byear{2023}).
\doiurl{10.1145/3617694.3623261}
\end{bchapter}
\endbibitem

\bibitem[\protect\citeauthoryear{Wiens et~al.}{2019}]{wiensNoHarmRoadmap2019}
\begin{barticle}
\bauthor{\bsnm{Wiens}, \binits{J.}},
\bauthor{\bsnm{Saria}, \binits{S.}},
\bauthor{\bsnm{Sendak}, \binits{M.}},
\bauthor{\bsnm{Ghassemi}, \binits{M.}},
\bauthor{\bsnm{Liu}, \binits{V.X.}},
\bauthor{\bsnm{Doshi-Velez}, \binits{F.}},
\bauthor{\bsnm{Jung}, \binits{K.}},
\bauthor{\bsnm{Heller}, \binits{K.}},
\bauthor{\bsnm{Kale}, \binits{D.}},
\bauthor{\bsnm{Saeed}, \binits{M.}},
\bauthor{\bsnm{Ossorio}, \binits{P.N.}},
\bauthor{\bsnm{Thadaney-Israni}, \binits{S.}},
\bauthor{\bsnm{Goldenberg}, \binits{A.}}:
\batitle{Do no harm: A roadmap for responsible machine learning for health care}.
\bjtitle{Nature Medicine}
\bvolume{25}(\bissue{9}),
\bfpage{1337}--\blpage{1340}
(\byear{2019})
\doiurl{10.1038/s41591-019-0548-6}
\end{barticle}
\endbibitem

\bibitem[\protect\citeauthoryear{Yao et~al.}{2023}]{Yao2023}
\begin{botherref}
\oauthor{\bsnm{Yao}, \binits{S.}},
\oauthor{\bsnm{Yu}, \binits{D.}},
\oauthor{\bsnm{Zhao}, \binits{J.}},
\oauthor{\bsnm{Shafran}, \binits{I.}},
\oauthor{\bsnm{Griffiths}, \binits{T.L.}},
\oauthor{\bsnm{Cao}, \binits{Y.}},
\oauthor{\bsnm{Narasimhan}, \binits{K.}}:
Tree of thoughts: Deliberate problem solving with large language models.
arXiv
(2023).
\url{https://arxiv.org/abs/2305.10601}
\end{botherref}
\endbibitem

\bibitem[\protect\citeauthoryear{Long}{2023}]{Long2023}
\begin{botherref}
\oauthor{\bsnm{Long}, \binits{J.}}:
Large language model guided tree-of-thought.
arXiv
(2023).
\url{https://arxiv.org/abs/2305.08291}
\end{botherref}
\endbibitem

\bibitem[\protect\citeauthoryear{Wang et~al.}{2022}]{Wang2022}
\begin{botherref}
\oauthor{\bsnm{Wang}, \binits{X.}},
\oauthor{\bsnm{Wei}, \binits{J.}},
\oauthor{\bsnm{Schuurmans}, \binits{D.}},
\oauthor{\bsnm{Le}, \binits{Q.}},
\oauthor{\bsnm{Chi}, \binits{E.}},
\oauthor{\bsnm{Narang}, \binits{S.}},
\oauthor{\bsnm{Chowdhery}, \binits{A.}},
\oauthor{\bsnm{Zhou}, \binits{D.}}:
Self-consistency improves chain of thought reasoning in language models.
arXiv
(2022).
\url{https://arxiv.org/abs/2203.11171}
\end{botherref}
\endbibitem

\bibitem[\protect\citeauthoryear{Nori et~al.}{2023}]{nori2023can}
\begin{botherref}
\oauthor{\bsnm{Nori}, \binits{H.}},
\oauthor{\bsnm{Lee}, \binits{Y.T.}},
\oauthor{\bsnm{Zhang}, \binits{S.}},
\oauthor{\bsnm{Carignan}, \binits{D.}},
\oauthor{\bsnm{Edgar}, \binits{R.}},
\oauthor{\bsnm{Fusi}, \binits{N.}},
\oauthor{\bsnm{King}, \binits{N.}},
\oauthor{\bsnm{Larson}, \binits{J.}},
\oauthor{\bsnm{Li}, \binits{Y.}},
\oauthor{\bsnm{Liu}, \binits{W.}},
\oauthor{\bsnm{Luo}, \binits{R.}},
\oauthor{\bsnm{McKinney}, \binits{S.M.}},
\oauthor{\bsnm{Ness}, \binits{R.O.}},
\oauthor{\bsnm{Poon}, \binits{H.}},
\oauthor{\bsnm{Qin}, \binits{T.}},
\oauthor{\bsnm{Usuyama}, \binits{N.}},
\oauthor{\bsnm{White}, \binits{C.}},
\oauthor{\bsnm{Horvitz}, \binits{E.}}:
Can generalist foundation models outcompete special-purpose tuning? {C}ase study in medicine.
arXiv
(2023).
\url{https://arxiv.org/abs/2311.16452}
\end{botherref}
\endbibitem

\bibitem[\protect\citeauthoryear{Bhatia}{2023}]{Bhatia2023}
\begin{barticle}
\bauthor{\bsnm{Bhatia}, \binits{S.}}:
\batitle{Inductive reasoning in minds and machines.}
\bjtitle{Psychological Review}
(\byear{2023})
\doiurl{10.1037/rev0000446}
\end{barticle}
\endbibitem

\bibitem[\protect\citeauthoryear{Zhang et~al.}{2022}]{zhang2022automatic}
\begin{botherref}
\oauthor{\bsnm{Zhang}, \binits{Z.}},
\oauthor{\bsnm{Zhang}, \binits{A.}},
\oauthor{\bsnm{Li}, \binits{M.}},
\oauthor{\bsnm{Smola}, \binits{A.}}:
Automatic chain of thought prompting in large language models.
arXiv
(2022).
\url{https://arxiv.org/abs/2210.03493}
\end{botherref}
\endbibitem

\bibitem[\protect\citeauthoryear{Fu et~al.}{2022}]{fu2022complexity}
\begin{botherref}
\oauthor{\bsnm{Fu}, \binits{Y.}},
\oauthor{\bsnm{Peng}, \binits{H.}},
\oauthor{\bsnm{Sabharwal}, \binits{A.}},
\oauthor{\bsnm{Clark}, \binits{P.}},
\oauthor{\bsnm{Khot}, \binits{T.}}:
Complexity-based prompting for multi-step reasoning.
arXiv
(2022).
\url{https://arxiv.org/abs/2210.00720}
\end{botherref}
\endbibitem

\bibitem[\protect\citeauthoryear{Salewski et~al.}{2023}]{Salewski2023}
\begin{bchapter}
\bauthor{\bsnm{Salewski}, \binits{L.}},
\bauthor{\bsnm{Alaniz}, \binits{S.}},
\bauthor{\bsnm{Rio-Torto}, \binits{I.}},
\bauthor{\bsnm{Schulz}, \binits{E.}},
\bauthor{\bsnm{Akata}, \binits{Z.}}:
\bctitle{In-context impersonation reveals large language models' strengths and biases}.
In: \beditor{\bsnm{Oh}, \binits{A.}},
\beditor{\bsnm{Naumann}, \binits{T.}},
\beditor{\bsnm{Globerson}, \binits{A.}},
\beditor{\bsnm{Saenko}, \binits{K.}},
\beditor{\bsnm{Hardt}, \binits{M.}},
\beditor{\bsnm{Levine}, \binits{S.}} (eds.)
\bbtitle{Advances in Neural Information Processing Systems (NeurIPS 2023)},
vol. \bseriesno{36},
pp. \bfpage{72044}--\blpage{72057}.
\bpublisher{Curran Associates, Inc.},
\blocation{New Orleans, LA}
(\byear{2023})
\end{bchapter}
\endbibitem

\bibitem[\protect\citeauthoryear{Brown et~al.}{2020}]{brown2020language}
\begin{botherref}
\oauthor{\bsnm{Brown}, \binits{T.}},
\oauthor{\bsnm{Mann}, \binits{B.}},
\oauthor{\bsnm{Ryder}, \binits{N.}},
\oauthor{\bsnm{Subbiah}, \binits{M.}},
\oauthor{\bsnm{Kaplan}, \binits{J.D.}},
\oauthor{\bsnm{Dhariwal}, \binits{P.}},
\oauthor{\bsnm{Neelakantan}, \binits{A.}},
\oauthor{\bsnm{Shyam}, \binits{P.}},
\oauthor{\bsnm{Sastry}, \binits{G.}},
\oauthor{\bsnm{Askell}, \binits{A.}},
\oauthor{\bsnm{Agarwal}, \binits{S.}},
\oauthor{\bsnm{Herbert-Voss}, \binits{A.}},
\oauthor{\bsnm{Krueger}, \binits{G.}},
\oauthor{\bsnm{Henighan}, \binits{T.}},
\oauthor{\bsnm{Child}, \binits{R.}},
\oauthor{\bsnm{Ramesh}, \binits{A.}},
\oauthor{\bsnm{Ziegler}, \binits{D.M.}},
\oauthor{\bsnm{Wu}, \binits{J.}},
\oauthor{\bsnm{Winter}, \binits{C.}},
\oauthor{\bsnm{Hesse}, \binits{C.}},
\oauthor{\bsnm{Chen}, \binits{M.}},
\oauthor{\bsnm{Sigler}, \binits{E.}},
\oauthor{\bsnm{Litwin}, \binits{M.}},
\oauthor{\bsnm{Gray}, \binits{S.}},
\oauthor{\bsnm{Chess}, \binits{B.}},
\oauthor{\bsnm{Clark}, \binits{J.}},
\oauthor{\bsnm{Berner}, \binits{C.}},
\oauthor{\bsnm{McCandlish}, \binits{S.}},
\oauthor{\bsnm{Radford}, \binits{A.}},
\oauthor{\bsnm{Sutskever}, \binits{I.}},
\oauthor{\bsnm{Amodei}, \binits{D.}}:
Language models are few-shot learners.
arXiv
(2020).
\url{https://arxiv.org/abs/2005.14165}
\end{botherref}
\endbibitem

\bibitem[\protect\citeauthoryear{Gu et~al.}{2020}]{pubmedbert}
\begin{botherref}
\oauthor{\bsnm{Gu}, \binits{Y.}},
\oauthor{\bsnm{Tinn}, \binits{R.}},
\oauthor{\bsnm{Cheng}, \binits{H.}},
\oauthor{\bsnm{Lucas}, \binits{M.}},
\oauthor{\bsnm{Usuyama}, \binits{N.}},
\oauthor{\bsnm{Liu}, \binits{X.}},
\oauthor{\bsnm{Naumann}, \binits{T.}},
\oauthor{\bsnm{Gao}, \binits{J.}},
\oauthor{\bsnm{Poon}, \binits{H.}}:
Domain-specific language model pretraining for biomedical natural language processing.
arXiv
(2020).
\url{https://arxiv.org/abs/2007.15779}
\end{botherref}
\endbibitem

\bibitem[\protect\citeauthoryear{Deka et~al.}{2022}]{deka2022improved}
\begin{barticle}
\bauthor{\bsnm{Deka}, \binits{P.}},
\bauthor{\bsnm{Jurek-Loughrey}, \binits{A.}},
\bauthor{\bsnm{Deepak}, \binits{P.}}:
\batitle{Improved methods to aid unsupervised evidence-based fact checking for online health news}.
\bjtitle{Journal of Data Intelligence}
\bvolume{3}(\bissue{4}),
\bfpage{474}--\blpage{504}
(\byear{2022})
\doiurl{10.26421/JDI3.4-5}
\end{barticle}
\endbibitem

\bibitem[\protect\citeauthoryear{Holcombe et~al.}{2021}]{holcombe2020documenting}
\begin{barticle}
\bauthor{\bsnm{Holcombe}, \binits{A.O.}},
\bauthor{\bsnm{Kovacs}, \binits{M.}},
\bauthor{\bsnm{Aust}, \binits{F.}},
\bauthor{\bsnm{Aczel}, \binits{B.}}:
\batitle{Documenting contributions to scholarly articles using credit and tenzing}.
\bjtitle{PLOS ONE}
\bvolume{15}(\bissue{12}),
\bfpage{1}--\blpage{11}
(\byear{2021})
\doiurl{10.1371/journal.pone.0244611}
\end{barticle}
\endbibitem

\end{thebibliography}

\end{document}